\documentclass{article} % For LaTeX2e
\usepackage{iclr2026_conference,times}
\iclrfinalcopy
\usepackage{booktabs}   % 三线表
\usepackage{multirow}   % 合并单元格

\usepackage{amsmath,amsfonts,bm}

\def\eqref#1{equation~\ref{#1}}
\def\1{\bm{1}}

\DeclareMathAlphabet{\mathsfit}{\encodingdefault}{\sfdefault}{m}{sl}
\SetMathAlphabet{\mathsfit}{bold}{\encodingdefault}{\sfdefault}{bx}{n}

\usepackage{booktabs}
\usepackage{multirow}
\usepackage{tabularx}
\usepackage{wrapfig}
\usepackage{enumitem}
\usepackage{graphicx}
\usepackage{colortbl} 
\usepackage{CJK}
\usepackage{hyperref}
\usepackage{svg}
\usepackage{url}
\usepackage{listings}
\usepackage[most]{tcolorbox}
\usepackage{markdown}
\usepackage{xcolor}  
\usepackage{fontawesome5}
\usepackage[utf8]{inputenc}
\usepackage[most]{tcolorbox}  
\usepackage{bbm}
\newcommand{\amapagent}{Amap Agent }

\newcommand{\eg}{e.g.}

\usepackage{xspace}
\newcommand{\model}{STAgent\xspace}

\definecolor{mycolor}{HTML}{eff6fd}
\definecolor{color_amapblue}{HTML}{3472ed}
\definecolor{color_amaplightblue}{HTML}{87e1f9}
\definecolor{color_amapgreen}{HTML}{c1fca0}
\definecolor{color_amapgrey}{HTML}{e5eff0}

\title{AMAP Agentic Planning Technical Report}

\makeatletter
\def\@maketitle{%
  \newpage
  \null
  \vskip 2em%
  \begin{center}% % 关键：确保标题在center环境中
    {\LARGE \@title \par}%
    \vskip 1.5em%
    {\large
      \begin{tabular}[t]{c}%
        \centering \@author
      \end{tabular}\par}%
    \vskip 1em%
  \end{center}%
  \par
  \vskip 1.5em
}
\makeatother

\author{
  \centerline{AMAP AI Agent LLM Team}
}

\begin{document}

\maketitle

\begin{abstract}
We present \model, an agentic large language model tailored for spatio-temporal understanding, designed to solve complex tasks such as constrained point-of-interest discovery and itinerary planning. \model is a specialized model capable of interacting with ten distinct tools within spatio-temporal scenarios, enabling it to explore, verify, and refine intermediate steps during complex reasoning. Notably, \model effectively preserves its general capabilities. We empower \model with these capabilities through three key contributions: (1) a stable tool environment that supports over ten domain-specific tools, enabling asynchronous rollout and training; (2) a hierarchical data curation framework that identifies high-quality data like a needle in a haystack, curating high-quality queries by retaining less than 1\% of the raw data, emphasizing both diversity and difficulty; and (3) a cascaded training recipe that starts with a seed SFT stage acting as a guardian to measure query difficulty, followed by a second SFT stage fine-tuned on queries with high certainty, and an ultimate RL stage that leverages data of low certainty. Initialized with Qwen3-30B-A3B-2507 to establish a strong SFT foundation and leverage insights into sample difficulty, \model yields promising performance on TravelBench while maintaining its general capabilities across a wide range of general benchmarks, thereby demonstrating the effectiveness of our proposed agentic model.
\end{abstract}

\section{Introduction}
The past year has witnessed the development of Large Language Models (LLMs) incorporating tool invocation for complex task reasoning, significantly pushing the frontier of general intelligence~\cite{kimi_k2,glm_4.5,deepseek_v32,qwen3}. Tool-integrated reasoning (TIR)~\cite{qu2025tool} empowers LLMs with the capability to interact with the tool environment, allowing the model to determine the next move based on feedback from the tools~\cite{tir_tora,tir_search,arpo,rstar2agent}. Crucially, existing TIR efforts mostly focus on scenarios like mathematical reasoning and code testing~\cite{agentic_rl_survey,TIR_understand}, while solutions for more practical real-world settings remain lacking.

\begin{figure}[h]
  \centering
  \includegraphics[width=0.98\textwidth]{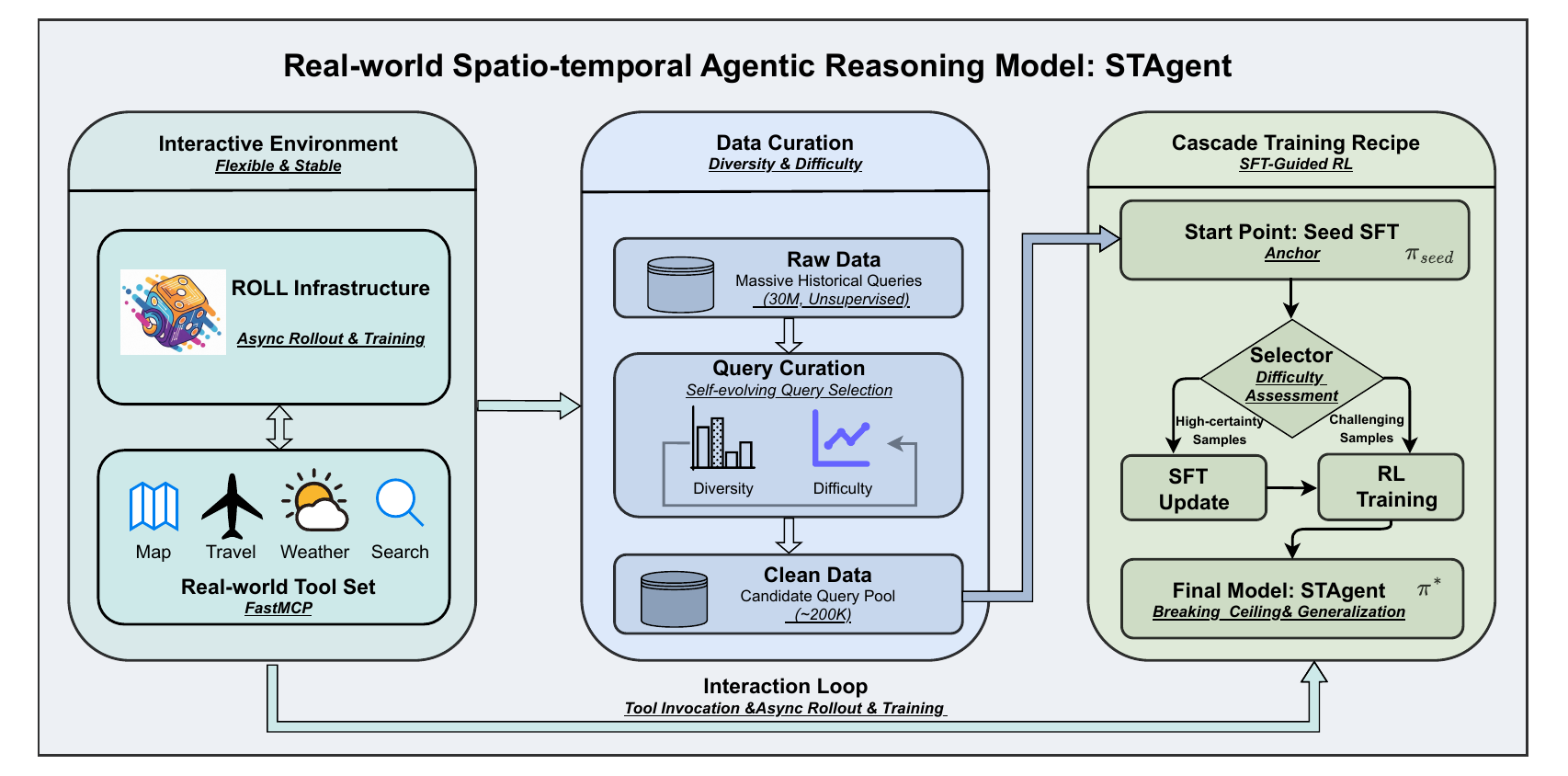}
  \caption{The overall framework of \model. It presents a comprehensive pipeline designed for real-world spatio-temporal reasoning. The framework consists of three key phases: (1) \textbf{Robust Interactive Environment}, supported by the ROLL infrastructure and FastMCP protocol to enable efficient, asynchronous tool-integrated reasoning. (2) \textbf{High-Quality Data Construction}, which utilizes a self-evolving selection framework to filter diverse and challenging queries from massive unsupervised data; (3) \textbf{Cascade Training Recipe}, an SFT-Guided RL paradigm that categorizes samples by difficulty to synergize supervised fine-tuning with reinforcement learning.}
\label{fig:framework}
\end{figure}

\begin{figure}
    \centering
    \includegraphics[width=1.0\linewidth]{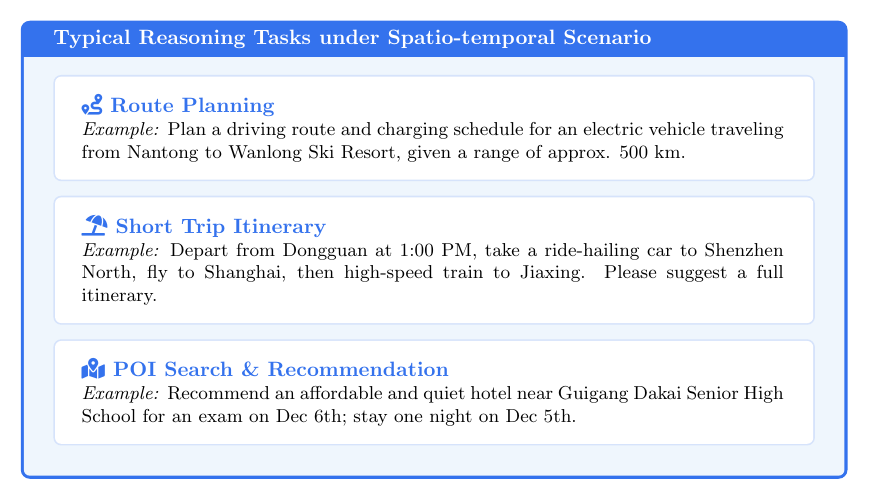}
    \vspace{-15pt} 
    \caption{Typical reasoning tasks under spatio-temporal scenario.}
    \label{fig:examples.}
\end{figure}

Real-world reasoning tasks can be categorized based on cognitive cost and processing speed into System 1 and System 2 modes~\cite{li2025system}: the former is rapid, whereas the latter necessitates extensive and complex deliberation. Reasoning tasks in spatio-temporal scenarios~\cite{zheng2025spatio,stma} represent typical System 2 scenarios. As depicted in Figure~\ref{fig:examples.}, such complex tasks involve identifying locations, designing driving routes, or planning travel itineraries subject to numerous constraints~\cite{ning2025deeptravel,xie2024travelplanner}, necessitating the coordination of heterogeneous external tools for resolution~\cite{agentic_rl_survey,toucan}. Consequently, TIR interleaves thought generation with tool execution, empowers the model to verify intermediate steps and dynamically adjust its planning trajectory based on observation feedback, and exhibits an inherent advantage in addressing these real-world tasks.

However, unlike general reasoning tasks such as mathematics~\cite{rstar2agent} or coding~\cite{glm_4.5}, addressing these real-world tasks inherently involves several challenges. \textbf{First}, how can a flexible and stable reasoning environment be constructed? On one hand, a feasible environment requires a toolset capable of handling massive concurrent tool call requests, which will be invoked by both offline data curation and online reinforcement learning (RL)~\cite{agentic_limits}. On the other hand, during training, maintaining effective synchronization between tool calls and trajectory rollouts, and guaranteeing that the model receives accurate reward signals, constitute the cornerstone of effective TIR for such complex tasks. \textbf{Second}, how can high-quality training data be curated? In real-world spatio-temporal scenarios, massive real-world queries are sent by users and stored in databases; however, this data is unsupervised and lacks necessary knowledge, \eg, the category and difficulty of each query~\cite{demystify_agentic_rl}, making the model unaware of the optimization direction~\cite{difficulty_know}. Therefore, it is critical to construct a query taxonomy to facilitate high-quality data selection for model optimization. \textbf{Third}, how should effective training for real-world TIR be conducted? Existing TIR efforts mostly focus on the adaptation between algorithms and tools. For instance, monitoring interleaved tool call entropy changes to guide further token rollout~\cite{arpo} or increasing rollout batch sizes to mitigate tool noise~\cite{rstar2agent}. Nevertheless, the uncertain environment and diverse tasks in real-world scenarios pose extra challenges when applying these methods. Therefore, deriving a more tailored training recipe is key to elevating the upper bound of TIR performance.

In this work, we propose \model, the pioneering agentic model designed for real-world spatio-temporal TIR reasoning. We develop a comprehensive pipeline encompassing a Interactive Environment, High-Quality Data Curation, and a Cascade Training Recipe, as shown in Figure~\ref{fig:framework}. Specifically, \model features three primary aspects. For environment and infrastructure, we established a robust reinforcement training environment supporting ten domain-specific tools across four categories, including map, travel, weather, and information retrieval tools. On the one hand, we encapsulated these tools using FastMCP~\footnote{\url{https://github.com/jlowin/fastmcp}}, standardizing parameter formats and invocation protocols, which significantly facilitates future tool modification. On the other hand, we collaborated with the ROLL~\cite{roll}~\footnote{\url{https://github.com/alibaba/ROLL}} team to optimize the RL training infrastructure, offering two core features: asynchronous rollout and training. Compared to the popular open-source framework Verl~\footnote{\url{https://github.com/volcengine/verl}}, ROLL yields an 80\% improvement in training efficiency. Furthermore, we designed a meticulous query selection framework to hierarchically extract high-quality queries from massive historical candidates. We focus on query diversity and difficulty, deriving a self-evolving query selection framework to filter approximately 200,000 queries from an original historical dataset of over 30 million, which serves as the candidate query pool for subsequent SFT and RL training. Lastly, we designed a cascaded training paradigm—specifically an SFT-Guided RL approach—to ensure continuous improvement in model capability. By training a Seed SFT model to serve as an evaluator, we assess the difficulty of the query pool to categorize queries for subsequent SFT updates and RL training. Specifically, SFT is refined with samples of higher certainty, while RL targets more challenging samples from the updated SFT model. This paradigm significantly enhances the generalization of the SFT model while enabling the RL model to push beyond performance ceilings.

The final model, \model, was built upon Qwen3-30B-A3B-2507~\cite{qwen3}. Notably, during the SFT stage, \model incorporated only a minimal amount of general instruction-following data~\cite{toucan} to enhance tool invocation capabilities, with the vast majority consisting of domain-specific data. Remarkably, as a specialized model, \model demonstrates significant advantages on TravelBench~\cite{Cheng2025TravelBenchAR} compared to models with larger parameter sizes. Furthermore, despite not being specifically tuned for the general domain, \model achieved improvements on numerous general-domain benchmarks, demonstrating its strong generalization capability.

\section{Methodology}
\subsection{Overview}
We formally present \model in this section, focusing on three key components:
\begin{itemize}
\item Tool environment construction. We introduce details of the in-domain tools used in Section~\ref{sec:env}.
\item High-quality prompt curation. We present the hierarchical prompt curation pipeline in Section~\ref{sec:prompt}, including the derivation of a prompt taxonomy, large-scale prompt annotation, and difficulty measurement.
\item Cascaded agentic post-training recipe. We present the post-training procedure in Section~\ref{sec:reward}, Section~\ref{sec:sft}, and Section~\ref{sec:rl}, which corresponds to reward design, agentic SFT, and SFT-guided RL training, respectively.
\end{itemize}

\subsection{Environment Construction} \label{sec:env}
To enable \model to interact with real-world spatio-temporal services in a controlled and reproducible manner, we developed a high-fidelity sandbox environment built upon FastMCP. This environment serves as the bridge between the agent's reasoning capabilities and the underlying spatio-temporal APIs, providing a standardized interface for tool invocation during both training and evaluation. To reduce API latency and costs during large-scale RL training, we implemented a tool-level LRU caching mechanism with parameter normalization to maximize cache hit rates.

Our tool library comprises \textbf{10 specialized tools} spanning four functional categories, designed to cover the full spectrum of spatio-temporal user needs identified in our Intent Taxonomy (Section \ref{sec:prompt}). All tool outputs are post-processed into structured natural language to facilitate the agent's comprehension and reduce hallucination risks. We describe the summary of the tool definition in Table~\ref{tab:tool_library}, more tool details can be found in Appendix~\ref{app:tools}.
\begin{table}[htbp]
\centering
\caption{Summary of the tool library for the \amapagent sandbox environment.}
\label{tab:tool_library}
\begin{tabular}{@{}lll@{}}
\toprule
\textbf{Category} & \textbf{Tool} & \textbf{Description} \\
\midrule
\multirow{5}{*}{Map \& Navigation}
& map\_search\_places & Search POIs by keyword, location, or region \\
& map\_compute\_routes & Route planning supporting multiple transport modes \\
& map\_search\_along\_route & Find POIs along a route corridor \\
& map\_search\_central\_places & Locate optimal meeting points for multiple origins \\
& map\_search\_ranking\_list & Query curated ranking lists \\
\midrule
\multirow{2}{*}{Travel}
& travel\_search\_flights & Search flights with optional multi-day range \\
& travel\_search\_trains & Query train schedules and fares \\
\midrule
\multirow{2}{*}{Weather}
& weather\_current\_conditions & Get real-time weather and AQI \\
& weather\_forecast\_days & Retrieve multi-day forecasts \\
\midrule
Information & web\_search & Open-domain web search \\
\bottomrule
\end{tabular}
\end{table}

\subsection{Prompt Curation} \label{sec:prompt} 

To endow the \model with comprehensive spatio-temporal reasoning capabilities, we synthesized a high-fidelity instruction dataset grounded in large-scale, real-world user behaviors. This dataset spans the full spectrum of user needs, covering atomic queries such as POI retrieval as well as composite, multi-constraint tasks like intricate itinerary planning.

We leveraged anonymized online user logs spanning a three-month window as our primary data source, with a volume of 30 million. The critical challenge lies in distilling these noisy, unstructured interactions into a structured, high-diversity instruction dataset suitable for training a sophisticated agent. To achieve this, we constructed a hierarchical \textbf{Intent Taxonomy}. This taxonomy functions as a rigorous framework for precise annotation, quantitative distribution analysis, and controlled sampling, ensuring the dataset maximizes both \textbf{Task Type Diversity} (comprehensive intent coverage) and \textbf{Difficulty Diversity} (progression from elementary to complex reasoning).

\subsubsection{Seed-Driven Taxonomy Evolution} We propose a \textbf{Seed-Driven Evolutionary Framework} to construct a taxonomy that guarantees both completeness and orthogonality (i.e., non-overlapping, independent dimensions). Instead of relying on static classification, our approach initiates with a high-quality kernel of seed prompts and iteratively expands the domain coverage by synergizing the generative power of LLMs with rigorous human oversight. The process unfolds in the following phases:

\begin{figure}[h]
  \centering
  \vspace{-20pt}
  \includegraphics[width=1\textwidth]{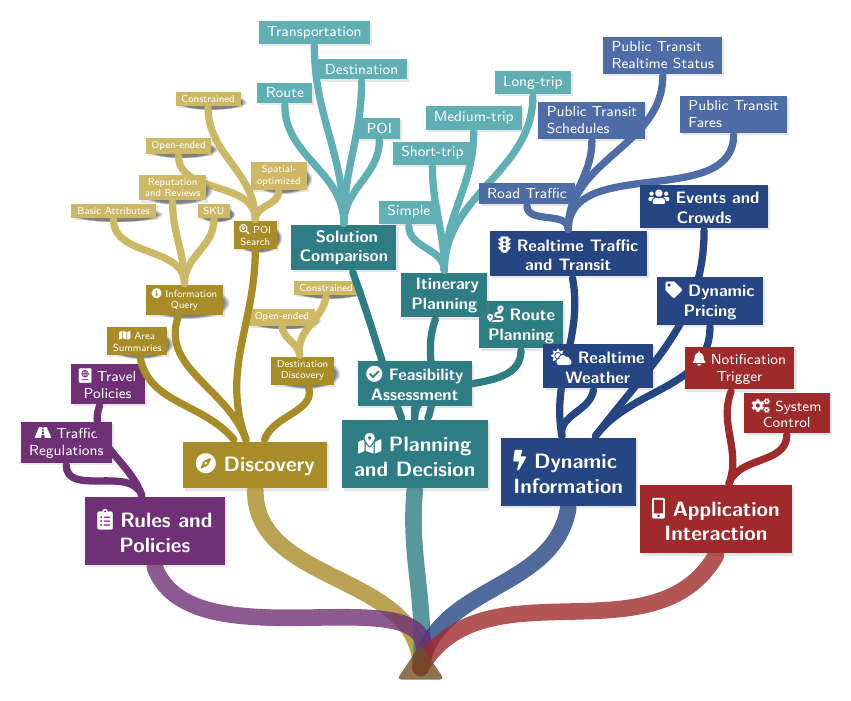}
  \caption{A visual taxonomy of our intent classification system. The hierarchical structure is organized into five primary categories: Rules and Policies, Discovery, Planning and Decision, Dynamic Information, and Application Interaction. The taxonomy further branches into 16 second-level categories and terminates in 30 fine-grained leaf nodes, capturing the multi-faceted complexity of real-world user queries in navigation and travel scenarios.}
  % \vspace{-20pt}  % ⭐ 增大负值，减少下方空白
  \label{fig:intent_taxonomy}
\end{figure}

{\textbf{Stage 1: Seed Initialization.}} We manually curate a small, high-variance set of $n$ seed prompts ($\mathcal{D}_{seed} \in \mathcal{D}_{pool}$) to represent the core diversity of the domain. Experts annotate these seeds with open-ended tags to capture abstract intent features. Each query is mapped into a tuple as:

$$\mathcal{S}_i = \langle q_i, \mathcal{T}_i \rangle, \text{ s.t. } \mathcal{T}_i = \{t_{i,1}, t_{i,2}, \dots, t_{i,k}\},$$

where $\mathcal{T}_i$ denotes the $k$ orthogonal or complementary intent nodes covered by instruction $q_i$.

{\textbf{Stage 2: LLM-Driven Category Induction.}} Using the annotated seeds, we prompt an LLM to induce orthogonal Level-1 categories, ensuring granular alignment across the domain.

{\textbf{Stage 3: Iterative Refinement Loop.}} To prevent hallucinations, we implement a strict check-update cycle. The LLM re-annotates the $\mathcal{D}_{seed}$ using the generated categories; human experts review these annotations to identify ambiguity or coverage gaps, feeding feedback back to the LLM for taxonomy refinement, executing a ``Tag-Feedback-Correction" cycle across several iterations:

$$\mathcal{T}^{(k+1)} = \text{Refine}\left( \text{Annotate}(\mathcal{D}_{seed}, \mathcal{T}^{(k)}), \text{Human\_Feedback} \right)$$

During this process, human experts identify ambiguity or coverage gaps in the LLM-generated categories, feeding critical insights back for taxonomy adjustment. The process terminates when $\mathcal{T}^{(k+1)} \approx \mathcal{T}^{(k)}$, signifying that the system has converged into a stable benchmark taxonomy $\mathcal{T}^*$.

{\textbf{Stage 4: Dynamic Taxonomy Expansion.}} To capture long-tail intents beyond the initial seeds, we enforce an ``Other" category at each node. By labeling a massive scale of raw logs and analyzing samples falling into the ``Other" category, we discover and merge emerging categories, allowing the taxonomy to evolve dynamically.

This mechanism transforms the taxonomy from a static tree into an evolving system capable of adapting to the open-world distribution of user queries. The resulting finalized Intent Taxonomy is visualized in Figure~\ref{fig:intent_taxonomy}.

\subsubsection{Annotation and Data Curation}

Building upon this stabilized and mutually orthogonal taxonomy, we deployed a high-capacity teacher LLM to execute large-scale intent classification on the structured log data. The specific prompt template used for this annotation is detailed in Appendix \ref{app:intent_classification_prompt}. To ensure the training data achieves both high information density and minimal redundancy, we implemented a rigorous two-phase process:

\textbf{Precise Multi-Dimensional Annotation.} We treat intent understanding as a multi-label classification task. Each user instruction is mapped to a composite label vector $\mathcal{V} = \langle I_{primary}, I_{secondary}, \mathcal{C}_{constraints} \rangle$.

\begin{itemize}
\item $I_{primary}$ and $I_{secondary}$ represent the leaf nodes in our fine-grained Intent Taxonomy.

\item $\mathcal{C}_{constraints}$ captures specific auxiliary dimensions (e.g., \textit{spatial\_range}, \textit{time\_budget}, \textit{vehicle\_type}).

\end{itemize}

This granular tagging captures the semantic nuance of complex queries, distinguishing between simple keyword searches and multi-constraint planning tasks.

\textbf{Controlled Sampling via Funnel Filtering.} 
Based on the annotation results, we applied a three-stage \textbf{Funnel Filtering Strategy} to construct the final curated dataset. This pipeline systematically eliminates redundancy at lexical, semantic, and geometric levels:
\begin{itemize}
    \item \textbf{Lexical Redundancy Elimination:} We applied global Locality-Sensitive Hashing \textbf{at corpus level} to efficiently remove near-duplicate strings and literal repetitions (``garbage data'') across all data, significantly reducing the initial volume.
    \item \textbf{Semantic Redundancy Elimination:} To ensure high intra-class variance, we partitioned the dataset into buckets based on the $\langle I_{primary}, I_{secondary} \rangle$ tuple. Within each bucket, we performed embedding-based similarity search to prune semantically redundant samples—defined as distinct phrasings that reside within a predefined distance threshold in the latent space. To handle the scale of our dataset, we integrated Faiss to accelerate this process; this transitioned the computational complexity from a quadratic $O(N^2)$ brute-force pairwise comparison to a more scalable sub-linear or near-linear search complexity, significantly reducing the preprocessing overhead.
    % Within each bucket, we utilized embedding-based similarity search to identify and prune semantically redundant samples (e.g., removing distinct phrasings that map to identical semantic vectors), ensuring high variance within each intent category.
    \item \textbf{Geometric Redundancy Elimination:} From the remaining pool, we employed the K-Center-Greedy algorithm to select the most representative samples. By maximizing the minimum distance between selected data points in the embedding space, this step preserves long-tail and corner-case queries that are critical for robust agent performance.
\end{itemize}

\begin{figure}[h]
  \centering
  \includegraphics[width=0.98\textwidth]{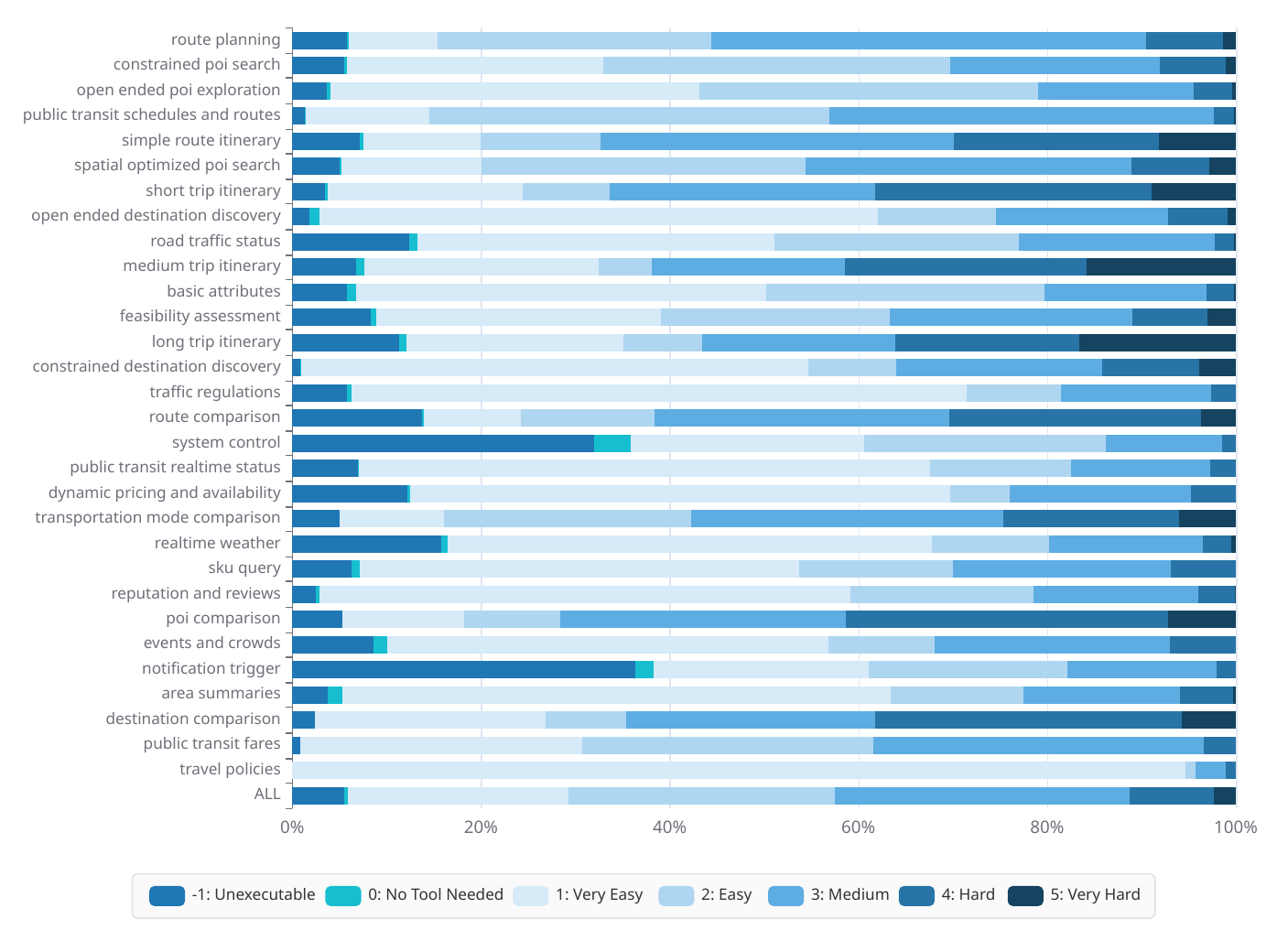}
  \caption{\textbf{Fine-grained Difficulty Distribution across 30 Geospatial Domains.} The visualization reveals the distinct complexity profiles of different tasks. While atomic queries (e.g., \textit{basic attributes}) cluster in low-difficulty regions (Score 1-2), composite tasks (e.g., \textit{long trip itinerary}) exhibit a significant proportion of high-complexity reasoning (Score 4-5). The visible segments of Score -1 and 0 (denoted as \textbf{Ext.1}) across domains like \textit{System Control} highlight our active sampling of boundary cases for hallucination mitigation.}
  \label{fig:prefilter-difficulty}
\end{figure}

\subsubsection{Difficulty and Diversity}
Real-world geospatial tasks vary significantly in cognitive load. To ensure the \model demonstrates robustness across this spectrum, we rigorously stratified our training data. Beyond mere coverage, the primary objective of this stratification is to enable Difficulty-based Curriculum Learning. In RL, a static data distribution often leads to training inefficiency: early-stage models facing excessively hard tasks yield zero rewards (vanishing gradients), while late-stage models facing trivial tasks yield constant perfect rewards (lack of variance). To address this, we employed an \textbf{Execution-Simulation Scoring Mechanism} to label data complexity, allowing us to dynamically align training data with model proficiency. This mechanism evaluates difficulty across three orthogonal dimensions:

\begin{itemize}
\item \textbf{Cognitive Load in Tool Selection:} Measures the ambiguity of intent mapping. It spans from \textit{Explicit Mapping} (Score 1-2, e.g., ``Navigates to X'') to \textit{Implicit Reasoning} (Score 4-5, e.g., ``Find a quiet place to read within a 20-minute drive,'' requiring abstract intent decomposition). 
\item \textbf{Execution Chain Depth:} Quantifies the logical complexity of the solution path, tracking the number and type of tool invocations and dependency depth (e.g., sequential vs. parallel execution). 
\item \textbf{Constraint Complexity:} Assesses the density of constraints (spatial, temporal, and preferential) that the agent must jointly optimize. 
\end{itemize}

To operationalize this mechanism at scale, we utilize a strong teacher LLM as an automated evaluator. This evaluator analyzes the structured user logs, assessing them against the three critical competencies to assign a scalar difficulty score $r \in \{-1, 0, 1, ..., 5\}$ to each sample. The specific prompt template used for this evaluation is detailed in Appendix \ref{app:difficulty_scoring_prompt}.

To validate the efficacy of this scoring mechanism, we visualize the distribution of annotated samples across 30 domains in Figure \ref{fig:prefilter-difficulty}. The distribution aligns intuitively with task semantics, providing strong support for our curriculum strategy:
\begin{itemize}\item \textbf{Spectrum Coverage for Curriculum Learning:} As shown in the ``ALL" bar at the bottom of Figure \ref{fig:prefilter-difficulty}, the aggregated dataset achieves a balanced stratification. This allows us to construct training batches that progressively shift from ``Atomic Operations" (Score 1-2) to ``Complex Reasoning" (Score 4-5) throughout the training lifecycle, preventing the reward collapse issues mentioned above.\item \textbf{Domain-Specific Complexity Profiles:} The scorer correctly identifies the inherent hardness of different intents. For instance, \textit{Long-trip Itinerary} is dominated by Score 4-5 samples (dark blue segments), reflecting its need for multi-constraint optimization. In contrast, \textit{Traffic Regulations} consists primarily of Score 1-2 samples, confirming the scorer's ability to distinguish between retrieval and reasoning tasks.
\end{itemize}

By curating batches based on these scores, we ensure the \model receives a consistent gradient signal throughout its training lifecycle. Furthermore, we place special emphasis on boundary conditions to enhance reliability:

\paragraph{Irrelevance and Hallucination Mitigation (Score -1/0).} A critical failure mode in agents is the tendency to hallucinate actions for unsolvable queries. To mitigate this, it is critical to train models to explicitly reject unanswerable queries or seek alternative solutions when desired tools are unavailable.

To achieve this, we constructed a dedicated \textbf{Irrelevance Dataset} (denoted as \textbf{Ext.1} in Figure \ref{fig:prefilter-difficulty}). By training on these negative samples, the \model learns to recognize the boundaries of its toolset, significantly reducing hallucination rates in open-world deployment.

\subsection{Reward Design}
\label{sec:reward}

To evaluate the quality of agent interactions, we employ a rubrics-as-reward~\citep{hashemi2024llm,gunjal2025rubrics} method to assess trajectories based on three core dimensions: \textit{Reasoning and Proactive Planning}, \textit{Information Fidelity and Integration}, and \textit{Presentation and Service Loop}. The scalar reward $R \in [0, 1]$ is derived from the following criteria:
\begin{itemize}
    \item \textbf{Dimension 1, Reasoning and Proactive Planning:} This dimension evaluates the agent's ability to formulate an economic and effective execution plan. A key metric is \textit{proactivity}: when faced with ambiguous or slightly incorrect user premises (e.g., mismatched location names), the agent is rewarded for correcting the error and actively attempting to solve the underlying intent, rather than passively rejecting the request. It also penalizes the inclusion of redundant parameters in tool calls.
    
    \item \textbf{Dimension 2, Information Fidelity and Integration:} This measures the accuracy with which the agent extracts and synthesizes information from tool outputs. We enforce a strict veto policy for hallucinations: any fabrication of factual data (e.g. time, price and distance) that cannot be grounded in the tool response results in an immediate reward of $0$. Conversely, the agent is rewarded for correctly identifying and rectifying factual errors in the user's query using tool evidence.
    
    \item \textbf{Dimension 3, Presentation and Service Loop:} This assesses whether the final response effectively closes the service loop. We prioritize responses that are structured, helpful, and provide actionable next steps. The agent is penalized for being overly conservative or terminating the service flow due to minor input errors.
\end{itemize}

Crucially, we recognize that different user intents require different capabilities. Therefore, we implement a \textbf{Dynamic Scoring Mechanism} where the evaluator autonomously determines the importance of each dimension based on the task type, rather than using fixed weights.

\paragraph{Dynamic Weight Determination.} For every query, the evaluator first analyzes the complexity of the user's request and categorizes it into one of three scenarios to assign specific weights ($w$):
\begin{itemize}
    \item \textbf{Scenario A: Complex Planning} (e.g., "Plan a 3-day itinerary"). The system prioritizes logical coherence and error handling. 
    \textit{Reference Weights:} $w_{\text{reas}} \approx 0.6, w_{\text{info}} \approx 0.3, w_{\text{pres}} \approx 0.1$.
    \item \textbf{Scenario B: Information Retrieval} (e.g., "What is the weather today?"). The system prioritizes factual accuracy and data extraction. 
    \textit{Reference Weights:} $w_{\text{reas}} \approx 0.2, w_{\text{info}} \approx 0.6, w_{\text{pres}} \approx 0.2$.
    \item \textbf{Scenario C: Consultation \& Explanation} (e.g., "Explain this policy"). The system prioritizes clarity and user interaction. 
    \textit{Reference Weights:} $w_{\text{reas}} \approx 0.3, w_{\text{info}} \approx 0.3, w_{\text{pres}} \approx 0.4$.
\end{itemize}

\paragraph{Score Aggregation and Hallucination Veto.}
Once the weights are established, the final reward $R$ is calculated using a weighted sum of the dimension ratings $s \in [0, 1]$. To ensure factual reliability, we introduce a \textbf{Hard Veto} mechanism for hallucinations. Let $\mathbb{1}_{H=0}$ be an indicator function where $H=1$ denotes the presence of hallucinated facts. The final reward is formulated as:
\begin{equation}
R = \mathbbm{1}_{H=0} \times \sum_{k \in \{\text{reas, info, pres}\}} w_k \cdot s_k.
\end{equation}
This formula ensures that any trajectory containing hallucinations is immediately penalized with a zero score, regardless of its reasoning or presentation quality.

\subsection{Supervised Fine-Tuning} \label{sec:sft}

Our Supervised Fine-Tuning (SFT) stage is designed to transition the base model from a general-purpose LLM into a specialized spatio-tempora agent. Rather than merely teaching the model to follow instructions, our primary objective is to develop three core agentic capabilities:

\textbf{Strategic Planning.} The ability to decompose abstract user queries (e.g., \textit{Plan a next-weekend trip}) into a logical chain of executable steps.

\textbf{Precise Tool Orchestration.} The capacity to orchastrate a series of tool calls leading to task fulfillment, ensuring syntactic correctness in parameter generation.

\textbf{Grounded Summarization.} The capability to synthesize final answer from heterogeneous tool outputs (e.g., \textit{maps, weather, flights}) into a coherent, well-organized response without hallucinating parameters not present in the observation.

\subsubsection{In-Domain Data Construction}

We construct our training data from the curated high-quality prompt pool derived in Section \ref{sec:prompt}. To address the long-tail distribution of real-world scenarios where frequent tasks like \textit{navigation} dominate while complex planning tasks are scarce, we employ a hybrid construction strategy:

\textbf{Offline Sampling with Strong LLMs.} For collected queries, we utilize a Strong LLM (DeepSeek-R1) to generate TIR trajectories. To ensure data quality, we generate $K=8$ candidate trajectories per query and employ a Verifier (Gemini-3-Pro-Preview) to score them based on the reward dimensions defined in Section \ref{sec:reward}. Only trajectories achieving perfect scores across all dimensions are retained.

\textbf{Synthetic Long-Tail Generation.} To bolster the model's performance on rare, complex tasks (e.g., multi-city itinerary planning with budget constraints), we employ In-Context Learning (ICL) to synthesize data. We sample complex tool combinations rarely seen in the existing data distribution and prompt a Strong LLM to synthesize user queries ($q_s$) that necessitate these specific tools in random orders. These synthetic queries are then validated through the offline sampling pipeline to ensure executability.

\subsubsection{Multi-Step Tool-Integrated Reasoning SFT}

We formalize the agent's interaction as a multi-step trajectory optimization problem. Unlike single-turn conversation, agentic reasoning requires the model to alternate between reasoning, tool invocation and tool observation processing. 

We model the probability of a trajectory $o$ as the product of conditional probabilities over tokens. The training objective minimizes the negative log-likelihood:

\begin{equation}
\mathcal{L}_{\mathrm{SFT}}(o_{i} \mid \theta) = \frac{1}{C} \sum_{t=1}^{|o_{i}|} \mathbb{I}(o_{i,t}) \log \pi_{\theta}(o_{i,t} \mid q_i)
\end{equation}

where $o_{i,t}$ denotes the $t$-th token of trajectory $o_{i}$. In practice, $q_s$ is optionally augmented with a corresponding user profile including user state (e.g., current location) and preferences (e.g., personal interests). $\mathbb{I}(o_{i,t})$ is a indicator function used for masking:
\begin{equation}
C = - \sum_{t=1}^{|o_{i}|} \mathbb{I}(o_{i,t})
\end{equation}
where text segments corresponding to tool observations do not contribute to the final loss calculation.

\subsubsection{Dynamic Capability-Aware Curriculum}
\begin{figure}[h]
  \centering
  \includegraphics[width=0.8\textwidth]{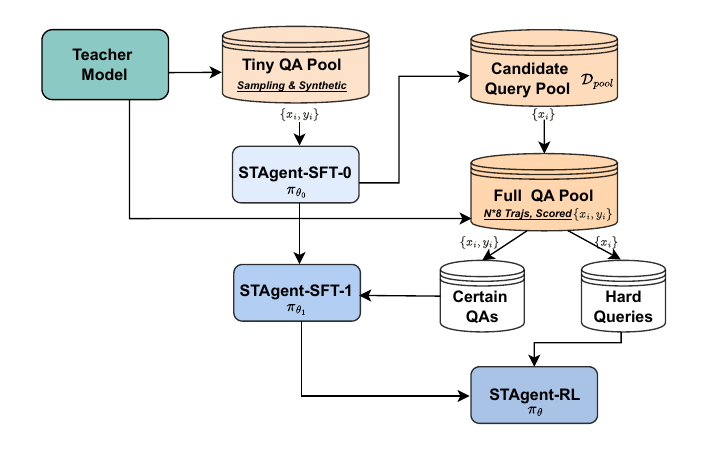}
  \caption{The training procedure in the SFT phase.}
  \label{fig:sft_stru}
\end{figure}

A core challenge in model training is determining which data points provide the highest information gain~\cite{difficulty_know}. Static difficulty metrics are insufficient because difficulty is inherently relative to the policy model's current parameterization. However, a task may be trivial for a strong LLM but lie outside the support of our policy model. Training on trivial samples yields vanishing gradients and increases the risk of overfitting, while training on impossible samples leads to high-bias updates and potential distribution collapse. In fact, samples that are trivial for a strong model but difficult for a weak model indicate a distribution gap between the two, and forcing the weak model to fit the distribution of the strong model in such cases can compromise generalization capability~\cite{weak2strong}. To bridge this gap, we define \textit{learnability} not as a static property of the prompt, but as the dynamic relationship between task difficulty and the policy's current capability. Learnable tasks are those that reside on the model's decision boundary—currently uncertain but solvable. Figure~\ref{fig:sft_stru} illustrates the training procedure of the SFT phase.

To address this, we introduce a \textbf{Dynamic Capability-Aware Curriculum} that actively selects data points based on the policy model's evolving capability distribution. This process consists of four phases:

\textbf{Phase 1: Policy Initialization.} 
We first establish a baseline capability to allow for meaningful self-assessment. We subsample a random 10\% Tiny Dataset from the curated prompt pool (denoted as $\mathcal{D}_{pool}$) and generate ground truth trajectories using the Strong LLM. We warm up our policy model by training on this subset to obtain the initial policy $\pi_{\theta_{0}}$. Specifically, for each prompt, we sample $K=8$ trajectories and employ a verifier to select the highest-scored generation as the ground truth.

\textbf{Phase 2: Distributional Capability Probing.} 
To construct a curriculum aligned with the policy's intrinsic capabilities, we perform a distributional probe of the initialized policy $\pi_{\theta_{0}}$ against the full unlabeled pool $\mathcal{D}_{pool}$. Rather than relying on external difficulty heuristics, we estimate the \textit{local solvability} of each task. For every query $q_i \in \mathcal{D}_{pool}$, we generate $K=8$ trajectories $\{o_{i,k}\}_{k=1}^K \sim \pi_{\theta_{0}}(\cdot|q_i)$ and use a verifier to evaluate their rewards $r_{i,k}$. This yields the mean the variance of the empirical reward distribution, $\hat{\mu}_i$ and $\hat{\sigma}^2_i$.

Here, $\hat{\sigma}^2_i$ serves as an approximate proxy for \textit{uncertainty}, where the policy parameters are unstable yet capable of generating high-reward solutions. Unlike random noise, when paired with a non-zero mean, this variance signifies the presence of a learnable signal within the policy's sampling space.

\textbf{Phase 3: Signal-to-Noise Filtration.} 
We formulate data selection as a signal maximization problem. Effective training samples must possess high gradient variance to drive learning while ensuring the gradient direction is grounded in the policy model's distribution. Based on the estimated statistics, we categorize the data into three regions:
\begin{itemize}
    \item \textbf{Trivial Region ($\hat{\mu} \approx 1, \hat{\sigma}^2 \to 0$):} Tasks where the policy has converged to high-quality solutions with high confidence. Training on these samples yields negligible gradients ($\nabla \mathcal{L} \approx 0$) and potentially leads to overfitting.
    \item \textbf{Noise Region ($\hat{\mu} \approx 0, \hat{\sigma}^2 \to 0$):} Tasks effectively outside the policy's current capability scope or containing noisy data. Training the model on these samples often leads to hallucination or negative transfer, as the ground truth lies too far outside the policy's effective support.
    \item \textbf{Learnable Region (High $\hat{\sigma}^2$, Non-zero $\hat{\mu}$):} Tasks maximizing training signals. These samples lie on the decision boundary where the policy is inconsistent but capable.
\end{itemize}
We retain only the Learnable Region. To quantify the learnability on a per-query basis, we introduce the \textit{Learnability Potential Score} $S_i = \hat{\sigma}^2_i \cdot \hat{\mu}_i$. This metric inherently prioritizes samples that are simultaneously uncertain (high $\hat{\sigma}^2$) and feasibly solvable (non-zero $\hat{\mu}$), maximizing the expected improvement in policy robustness. The scores are normalized between 0-1.

\textbf{Phase 4: Adaptive Trajectory Synthesis.} 
Having identified the high-value distribution, we employ an adaptive compute allocation strategy to construct the final SFT dataset. We treat the Strong LLM as an expensive oracle. We allocate the sampling budget $B_i$ proportional to the task's learnability score, such that $B_i \propto \text{rank}(S_i)$. Tasks with more uncertainty receive up to $K_{max}=8$ samples from the Strong LLM to maximize the probability of recovering a valid reasoning path, while easier tasks receive fewer calls. This ensures that the supervision signal is densest where the policy model is most uncertain, effectively correcting the policy's decision boundary with high-precision ground truth.

Finally, we aggregate the verified trajectories obtained from this phase. We empirically evaluate various data mixing ratios to determine the optimal training distribution, obtaining $\pi_{\theta_{1}}$, which serves as the backbone model for subsequent RL training.

\subsection{RL} \label{sec:rl}
The agentic post-training for \model follows the "SFT-Guided RL" training paradigm, with the RL policy model initialized from . This initialization provides the agent with foundational instruction-following capabilities prior to the exploration phase in the sandbox environment.

GRPO~\citep{shao2024deepseekmath} has become the de facto standard for reasoning tasks. Training is conducted within a high-fidelity sandbox environment (described in Section \ref{sec:env}), which simulates real-world online scenarios. This setup compels the agent to interact dynamically with the environment, specifically learning to verify and execute tool calls to resolve user queries effectively. The training objective seeks to maximize the expected reward while restricting the policy update to prevent significant deviations from the reference model. We formulate the GRPO objective for \model as follows:
\begin{equation}
\begin{split}
\mathcal{J}(\theta) = \mathbb{E}_{q \sim P(Q), \{o_i\}_{i=1}^G \sim \pi_{\theta_{\text{old}}}(O|q)} \Bigg[ \frac{1}{G} \sum_{i=1}^G \frac{1}{|o_i|} \sum_{t=1}^{|o_i|} \Bigg( \min \Bigg( \frac{\pi_\theta(o_{i,t} | q, o_{i,<t})}{\pi_{\theta_{\text{old}}}(o_{i,t} | q, o_{i,<t})} \hat{A}_i, \\
\text{clip} \left( \frac{\pi_\theta(o_{i,t} | q, o_{i,<t})}{\pi_{\theta_{\text{old}}}(o_{i,t} | q, o_{i,<t})}, 1-\epsilon, 1+\epsilon \right) \hat{A}_i \Bigg) - \beta \mathbb{D}_{KL} \Bigg) \Bigg],
\end{split}
\end{equation}
where $\pi_{\theta}$ and $\pi_{\theta_{\text{old}}}$ denote the current and old policies, respectively. 
The optimization process operates iteratively on batches of queries. For each query $q$, we sample a group of $G$ independent trajectories $\{o_1, o_2, \dots, o_G\}$ from the  policy $\pi_\theta$. Once the trajectories are fully rolled out, the reward function detailed in Section \ref{sec:reward} evaluates the quality of the interaction. The advantage $\hat{A}_i$ for the $i$-th trajectory is computed as:
$
\hat{A}_i = \frac{r_i - \text{mean}(\{r_1, \dots, r_G\})}{\text{std}(\{r_1, \dots, r_G\}) + \delta}.
$ $\mathbb{D}_{KL}$ refers to the KL divergence between the current policy $\pi_\theta$ and the reference policy $\pi_{\text{ref}}$. $\beta$ is the KL coefficient.

\model is built upon Qwen3-30B-A3B, which features a Mixture-of-Experts (MoE) architecture. In practice, we employ the GRPO variant, Group Sequence Policy Optimization (GSPO)~\citep{zheng2025group}, to stabilize training. Unlike the standard token-wise formulation, GSPO enforces a sequence-level optimization constraint by redefining the importance ratio. Specifically, it computes the ratio as the geometric mean of the likelihood ratios over the entire generated trajectory length :
\begin{equation}
s_i(\theta) = \exp \left( \frac{1}{|o_i|} \sum_{t=1}^{|o_i|} \log \frac{\pi_\theta(o_{i,t}|q, o_{i,<t})}{\pi_{\theta_{\text{old}}}(o_{i,t}|q, o_{i,<t})} \right).
\end{equation}

\section{Experiments}
\subsection{Experiment Setups}

Our evaluation protocol is designed to comprehensively assess \model across two primary dimensions: domain-specific expertise and general-purpose capabilities across a wide spectrum of tasks.

\subsubsection*{In-domain Evaluation}

To evaluate the \model's specialized performance in real-world and simulated travel scenarios, we conduct assessments in two environments:

\textbf{Indomain Online Evaluation.} To rapidly evaluate performance in real online environments, we extracted 1,000 high-quality queries covering five task types and seven difficulty levels (as detailed in Section \ref{sec:prompt}). For each query, we performed inference 8 times and calculated the average scores. We employ Gemini-3-flash-preview as the judge to compare the win rates between Amap Agent and baselines across different dimensions.

\textbf{Indomain Offline Evaluation.} We evaluate Amap Agent on TravelBench~\cite{Cheng2025TravelBenchAR}, a static sandbox environment containing multi-turn, single-turn, and unsolvable subsets. Following the default protocol, we run inference three times per query (temperature 0.7) and report the average results. GPT-4.1-0414 is used to simulate the user, and Gemini-3-flash-preview serves as the grading model.

\subsubsection*{General Capabilities Evaluation}

To comprehensively evaluate the general capabilities of our model, we conduct assessments across a diverse set of benchmarks.
Unless otherwise specified, we standardize the evaluation settings across all general benchmarks with a temperature of 0.6 and a maximum generation length of 32k tokens.
The benchmarks are categorized as follows:

\textbf{Tool Use \& Agentic Capabilities.} To test the model's proficiency in tool use and complex agentic interactions, we utilize ACEBench~\cite{acebench} and $\tau^{2}$-Bench~\cite{tau2bench}. Furthermore, we evaluate function-calling capabilities using the BFCL v3~\cite{bfcl}.

\textbf{Mathematical Reasoning.} We assess advanced mathematical problem-solving abilities using the AIME 24 and AIME 25~\cite{aime}, which serve as proxies for complex logical reasoning.

\textbf{Coding.} We employ LiveCodeBench~\cite{livecodebench} to evaluate coding capabilities on contamination-free problems. We evaluate on both the v5 (167 problems) and v6 (175 problems) subsets. Specifically, we set the temperature to 0.2, sample once per problem, and report the Pass@1 accuracy.

\textbf{General \& Alignment Tasks.} We evaluate general knowledge and language understanding using MMLU-Pro~\cite{mmlupro} and, specifically for Chinese proficiency, C-Eval~\cite{ceval}. To assess how well the model aligns with human preferences and instruction following, we employ ArenaHard-v2.0~\cite{arenahard} and IFEval~\cite{ifeval}, respectively.

\subsection{Main Results}

\begin{figure}[ht]
    \centering
    \includegraphics[width=0.85\linewidth]%{resources/model_comparison_indomain.pdf}
    {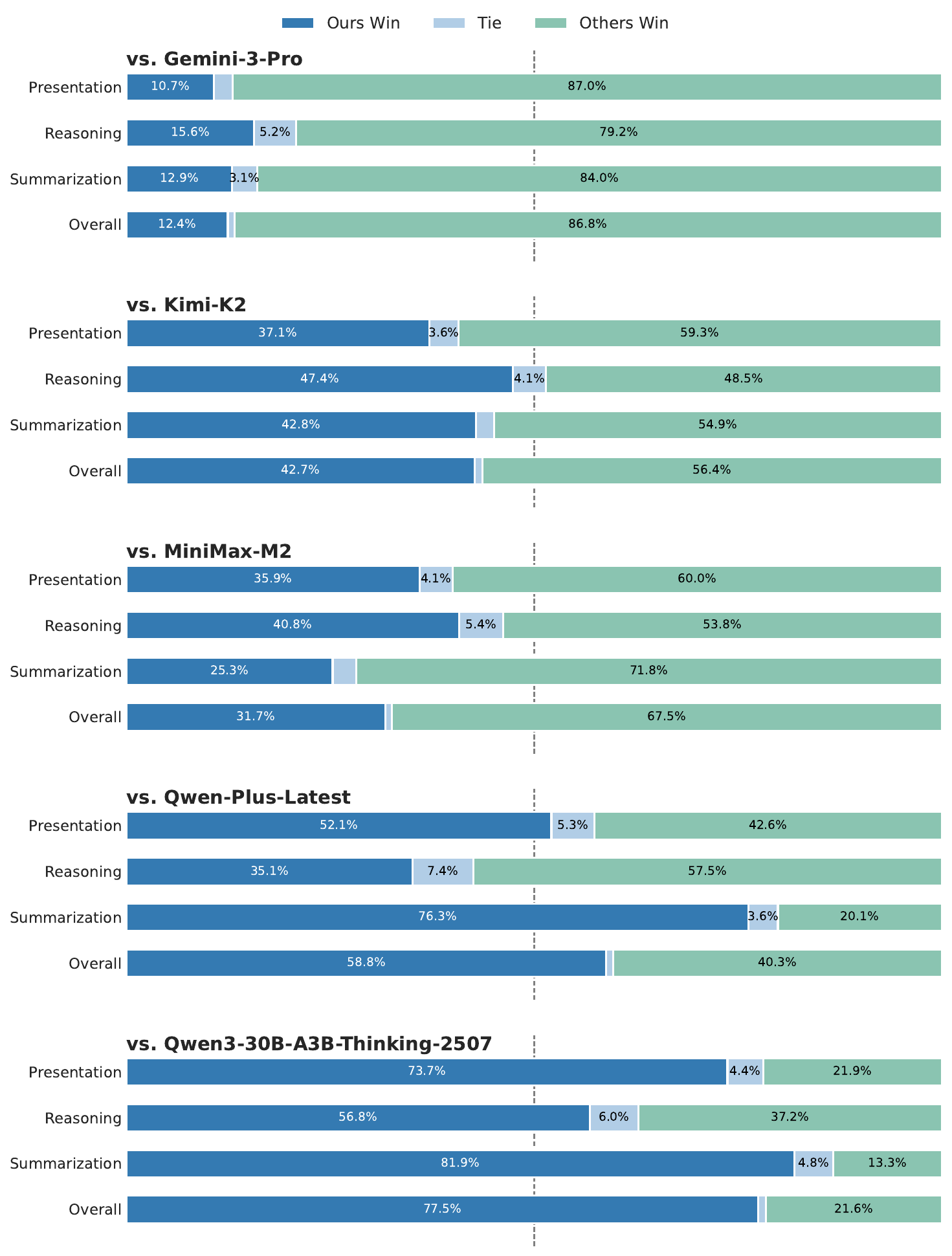}
    \caption{Amap indomain online benchmark evaluation.}
    \label{fig:win_rate}
\end{figure}

\paragraph{Amap Indomain online benchmark.} The experimental results, as illustrated in Figure \ref{fig:win_rate}, demonstrate the superior performance and robustness of \model. Specifically, Compared with the baseline model, Qwen3-30B-A3B-Thinking-2507, \model demonstrates substantial improvements across all three evaluation dimensions. Notably, in terms of contextual summarization/extraction (win rate: 81.9\%) and content presentation (win rate: 73.7\%), our model exhibits a robust capability to accurately synthesize preceding information and effectively present solutions that align with user requirements. When compared to Qwen-Plus-Latest, while \model shows a performance gap in reasoning and planning, it achieves a marginal lead in presentation and a significant advantage in the summarization dimension. Regarding other models such as MiniMax-M2, Kimi-K2, and Gemini-3-Pro-preview, a performance disparity persists across all three dimensions; we attribute this primarily to the constraints of model scale (30B), which limit the further enhancement of \model’s capabilities. Overall, these results validate the efficacy and promising potential of our proposed solution in strengthening tool invocation, response extraction, and summarization within map-based scenarios.

\paragraph{Amap Indomain offline benchmark.} 
As shown in Table~\ref{tab:travelbench}, our model trained on top of Qwen3-30B achieves consistent improvements across all three subtasks, with gains of \textbf{+11.7\%} (Multi-turn), \textbf{+5.8\%} (Single-turn), and \textbf{+26.1\%} (Unsolved) over the Qwen3-30B-thinking baseline. It also delivers a higher overall score (\textbf{70.3}) than substantially larger models such as DeepSeek R1 and Qwen3-235B-Instruct. Notably, our model attains the best performance on the Multi-turn subtask (\textbf{66.6}) and the second-best performance on the Single-turn subtask (\textbf{73.4}), which further supports the effectiveness of our training pipeline.

\paragraph{General domain evaluation.} The main experimental results, summarized in Table~\ref{tab:experimental_results_transposed}, shows several key observation. Firstly, \model achieves a significant performance leap on the private domain benchmark, marking a substantial improvement over its initialization model, Qwen3-30B-A3B-Thinking. Notably, \model surpasses larger-scale models, including Qwen3-235B-A22B-Thingking-2507 and DeepSeek-R1-0528, while comparable to Qwen3-235B-A22B-Instruct-2507. Secondly, \model shows a notable performance in Tool Use benchmarks, which suggests that models trained on our private domain tools possess a strong ability to generalize their tool-calling capabilities to other diverse functional domains. Finally, the model effectively maintains its high performance across public benchmarks for mathematics, coding, and general capabilities , proving that our domain-specific reinforcement learning process does not lead to a degradation of general-purpose performance or fundamental reasoning skills. Despite being a specialized model for the spatio-temporal domain, \model still achieves excellent performance in general domains while maintaining its strong in-domain capabilities, demonstrating the effectiveness of our training methodology.

% \begin{table}[]
% \centering
% \caption{Results on TravelBench, covering three subtasks: Multi-turn, Single-turn, and Unsolved.}
%   \label{tab:travelbench}
% \begin{tabular}{@{}ccccc@{}}
% \toprule
% Model                 & Multi-turn & Single-turn & Unsolved & Overall \\ \midrule
% Deepseek R1-0528      & 34.3      & 76.1       & 83.7    & 64.7   \\
% Qwen3-235B-A22B-Ins-2507   & 60.1      & 69.7       & 80.0       & \underline{69.9}   \\
% Qwen3-235B-A22B-Th-2507   & 56.6      & 73.2       & 51.7    & 60.5   \\
% Qwen3-30B-A3B-Th-2507    & 59.6      & 69.4       & 56.3    & 61.8   \\
% Qwen3-14B             & 47.0      & 57.0       & 54.0       & 52.7   \\
% Qwen3-4B              & 42.0      & 42.1       & 73.0       & 52.4   \\ \midrule
% \model & 66.6      & 73.4       & 71.0       & \textbf{70.3}   \\ \bottomrule
% \end{tabular}
% \end{table}

\begin{table}[ht]
\centering
\caption{Results on TravelBench, covering three subtasks: Multi-turn, Single-turn, and Unsolved.}
\label{tab:travelbench}
\resizebox{0.8\textwidth}{!}{
\begin{tabular}{@{}ccccc@{}}
\toprule
Model                 & Multi-turn & Single-turn & Unsolved & Overall \\ \midrule
Deepseek R1-0528      & 34.3      & 76.1       & 83.7    & 64.7   \\
Qwen3-235B-A22B-Ins-2507   & 60.1      & 69.7       & 80.0       & \underline{69.9}   \\
Qwen3-235B-A22B-Th-2507   & 56.6      & 73.2       & 51.7    & 60.5   \\
Qwen3-30B-A3B-Th-2507    & 59.6      & 69.4       & 56.3    & 61.8   \\
Qwen3-14B             & 47.0      & 57.0       & 54.0       & 52.7   \\
Qwen3-4B              & 42.0      & 42.1       & 73.0       & 52.4   \\ \midrule
\model & 66.6 & 73.4 & 71.0 & \textbf{70.3} \\
\bottomrule
\end{tabular}
}
\end{table}

\begin{table}[h]
\centering
\caption{Model performance evaluation across general and in-domain benchmarks.}
\label{tab:experimental_results_transposed}
\resizebox{\textwidth}{!}{
\begin{tabular}{@{}cccccccccc@{}}
\toprule
Domain                    & Benchmark        & \begin{tabular}[c]{@{}c@{}}Qwen3-4B-\\ Thinking-2507\end{tabular} & \begin{tabular}[c]{@{}c@{}}Qwen3\\ 14B\end{tabular} & \begin{tabular}[c]{@{}c@{}}Qwen3-30B-\\ A3B-2507\end{tabular} & \begin{tabular}[c]{@{}c@{}}Qwen3-235B-\\ A22B-Ins-2507\end{tabular} & \begin{tabular}[c]{@{}c@{}}Qwen3-235B-\\ A22B-Th-2507\end{tabular} & \begin{tabular}[c]{@{}c@{}}DeepSeek-\\ R1-0528\end{tabular} & STAgent & 
\begin{tabular}[c]{@{}c@{}}$\Delta$\\ (30B-A3B)\end{tabular}
 \\ \midrule
\multirow{3}{*}{Tool Use} & ACEBench         & 71.7                                                              & 69.8      & 75.7                                                          & 75.6                                                                & 75.7                                                               & -                                                           & 75.3    & -0.4  \\
                          & Tau2-Bench       & 46.2                                                              & 37.6      & 47.7                                                          & 52.4                                                                & 58.5                                                               & 52.7                                                        & 47.0      & -0.7  \\
                          & BFCL V3          & 71.2                                                              & 70.4      & 72.4                                                          & 70.9                                                                & 71.9                                                               & 63.8                                                        & 76.8    & 4.4   \\ \midrule
\multirow{2}{*}{Math}     & AIME 24          & 83.8                                                              & 79.3      & 91.3                                                          & 80.8                                                                & 93.8                                                               & 91.4                                                        & 90.2    & -0.9  \\
                          & AIME 25          & 81.3                                                              & 70.4      & 85.0                                                          & 70.3                                                                & 92.3                                                               & 87.5                                                        & 85.2    & 0.2   \\ \midrule
\multirow{2}{*}{Coding}   & LiveCodeBench-v5 & 61.7                                                              & 63.5      & 70.1                                                          & 57.5                                                                & 68.3                                                               & -                                                           & 70.7    & 0.6   \\
                          & LiveCodeBench-v6 & 55.2                                                              & 55.4      & 66.0                                                          & 51.8                                                                & 74.1                                                               & 73.3                                                        & 66.3    & 0.3   \\ \midrule
\multirow{4}{*}{General}  & ArenaHard-v2.0   & 34.9                                                              & 30.4      & 51.4                                                          & 79.2                                                                & 79.7                                                               & -                                                           & 46.4    & -5.0  \\
                          & IFEval           & 87.4                                                              & 85.4      & 88.9                                                          & 88.7                                                                & 87.8                                                               & 79.1                                                        & 87.1    & -1.8  \\
                          & MMLU-Pro         & 74.0                                                              & 77.4      & 80.9                                                          & 83.0                                                                & 84.4                                                               & 85.0                                                          & 80.5    & -0.4  \\
                          & C-Eval           & 72.3                                                              & 87.5      & 87.1                                                          & 90.7                                                                & 92.0                                                               & 91.5                                                        & 87.9    & 0.8   \\ \midrule
In-domain                 & TravelBench      & 60.8                                                              & 52.7      & 60.2                                                          & 69.9                                                                & 60.5                                                               & 64.7                                                        & 70.3    & 10.1  \\ \bottomrule
\end{tabular}
}
\end{table}

\section{Conclusion}
In this work, we present \model, a agentic model specifically designed to address the complex reasoning tasks within real-world spatio-temporal scenarios. A stable tool calling environment, high-quality data curation, and a difficulty-aware training recipe collectively contribute to the model's performance. Specifically, we constructed a calling environment that supports tools across 10 distinct domains, enabling stable and highly concurrent operations. Furthermore, we curated a set of high-quality candidate queries from a massive corpus of real-world historical queries, employing an exceptionally low filtering ratio. Finally, we designed an SFT-guided RL training strategy to ensure the model's capabilities continuously improve  throughout the training process. Empirical results demonstrate that \model, built on Qwen3-30B-A3B-thinking-2507, significantly outperforms models with larger parameter sizes on domain-specific benchmarks such as TravelBench, while maintaining strong generalization across general tasks. We believe this work not only provides a robust solution for spatio-temporal intelligence but also offers a scalable and effective paradigm for developing specialized agents in other complex, open-ended real-world environments.
\subsubsection*{Author Contributions}
\textbf{Project Leader}

Yulan Hu

\textbf{Core Contributors}

Xiangwen Zhang, Sheng Ouyang, Hao Yi, Lu Xu, Qinglin Lang, Lide Tan, Xiang Cheng

\textbf{Contributors}

Tianchen Ye, Zhicong Li, Ge Chen, Wenjin Yang, Zheng Pan, Shaopan Xiong, Siran Yang, Ju Huang, Yan Zhang, Jiamang Wang, Yong Liu, Yinfeng Huang, Ning Wang, Tucheng Lin, Xin Li, Ning Guo

\subsubsection*{Acknowledgments}
We extend our sincere gratitude to the ROLL~\cite{roll} team for their exceptional support on the training infrastructure. Their asynchronous rollout and training strategy enabled a nearly 80\% increase in training efficiency. We also thank Professor Yong Liu from the Gaoling School of Artificial Intelligence, Renmin University of China, for his insightful guidance on our SFT-guided RL training recipe.

\bibliography{iclr2026_conference}

@article{shao2024deepseekmath,
  title={Deepseekmath: Pushing the limits of mathematical reasoning in open language models},
  author={Shao, Zhihong and Wang, Peiyi and Zhu, Qihao and Xu, Runxin and Song, Junxiao and Bi, Xiao and Zhang, Haowei and Zhang, Mingchuan and Li, YK and Wu, Yang and others},
  journal={arXiv preprint arXiv:2402.03300},
  year={2024}
}

@article{zheng2025group,
  title={Group sequence policy optimization},
  author={Zheng, Chujie and Liu, Shixuan and Li, Mingze and Chen, Xiong-Hui and Yu, Bowen and Gao, Chang and Dang, Kai and Liu, Yuqiong and Men, Rui and Yang, An and others},
  journal={arXiv preprint arXiv:2507.18071},
  year={2025}
}

@article{glm_4.5,
  title={Glm-4.5: Agentic, reasoning, and coding (arc) foundation models},
  author={Zeng, Aohan and Lv, Xin and Zheng, Qinkai and Hou, Zhenyu and Chen, Bin and Xie, Chengxing and Wang, Cunxiang and Yin, Da and Zeng, Hao and Zhang, Jiajie and others},
  journal={arXiv preprint arXiv:2508.06471},
  year={2025}
}

@article{kimi_k2,
  title={Kimi k2: Open agentic intelligence},
  author={Team, Kimi and Bai, Yifan and Bao, Yiping and Chen, Guanduo and Chen, Jiahao and Chen, Ningxin and Chen, Ruijue and Chen, Yanru and Chen, Yuankun and Chen, Yutian and others},
  journal={arXiv preprint arXiv:2507.20534},
  year={2025}
}

@article{deepseek_v32,
  title={Deepseek-v3. 2: Pushing the frontier of open large language models},
  author={Liu, Aixin and Mei, Aoxue and Lin, Bangcai and Xue, Bing and Wang, Bingxuan and Xu, Bingzheng and Wu, Bochao and Zhang, Bowei and Lin, Chaofan and Dong, Chen and others},
  journal={arXiv preprint arXiv:2512.02556},
  year={2025}
}

@article{tir_tora,
  title={Tora: A tool-integrated reasoning agent for mathematical problem solving},
  author={Gou, Zhibin and Shao, Zhihong and Gong, Yeyun and Shen, Yelong and Yang, Yujiu and Huang, Minlie and Duan, Nan and Chen, Weizhu},
  journal={arXiv preprint arXiv:2309.17452},
  year={2023}
}

@article{tir_search,
  title={Learning to reason with search for llms via reinforcement learning},
  author={Chen, Mingyang and Sun, Linzhuang and Li, Tianpeng and Sun, Haoze and Zhou, Yijie and Zhu, Chenzheng and Wang, Haofen and Pan, Jeff Z and Zhang, Wen and Chen, Huajun and others},
  journal={arXiv preprint arXiv:2503.19470},
  year={2025}
}

@article{arpo,
  title={Agentic reinforced policy optimization},
  author={Dong, Guanting and Mao, Hangyu and Ma, Kai and Bao, Licheng and Chen, Yifei and Wang, Zhongyuan and Chen, Zhongxia and Du, Jiazhen and Wang, Huiyang and Zhang, Fuzheng and others},
  journal={arXiv preprint arXiv:2507.19849},
  year={2025}
}

@article{rstar2agent,
  title={rstar2-agent: Agentic reasoning technical report},
  author={Shang, Ning and Liu, Yifei and Zhu, Yi and Zhang, Li Lyna and Xu, Weijiang and Guan, Xinyu and Zhang, Buze and Dong, Bingcheng and Zhou, Xudong and Zhang, Bowen and others},
  journal={arXiv preprint arXiv:2508.20722},
  year={2025}
}

@article{li2025system,
  title={From system 1 to system 2: A survey of reasoning large language models},
  author={Li, Zhong-Zhi and Zhang, Duzhen and Zhang, Ming-Liang and Zhang, Jiaxin and Liu, Zengyan and Yao, Yuxuan and Xu, Haotian and Zheng, Junhao and Wang, Pei-Jie and Chen, Xiuyi and others},
  journal={arXiv preprint arXiv:2502.17419},
  year={2025}
}

@article{ning2025deeptravel,
  title={DeepTravel: An End-to-End Agentic Reinforcement Learning Framework for Autonomous Travel Planning Agents},
  author={Ning, Yansong and Liu, Rui and Wang, Jun and Chen, Kai and Li, Wei and Fang, Jun and Zheng, Kan and Tan, Naiqiang and Liu, Hao},
  journal={arXiv preprint arXiv:2509.21842},
  year={2025}
}

@article{zheng2025spatio,
  title={Spatio-temporal llm: Reasoning about environments and actions},
  author={Zheng, Haozhen and Tian, Beitong and Wu, Mingyuan and Tang, Zhenggang and Nahrstedt, Klara and Schwing, Alex},
  journal={arXiv preprint arXiv:2507.05258},
  year={2025}
}

@article{xie2024travelplanner,
  title={Travelplanner: A benchmark for real-world planning with language agents},
  author={Xie, Jian and Zhang, Kai and Chen, Jiangjie and Zhu, Tinghui and Lou, Renze and Tian, Yuandong and Xiao, Yanghua and Su, Yu},
  journal={arXiv preprint arXiv:2402.01622},
  year={2024}
}

@article{toucan,
  title={Toucan: Synthesizing 1.5 m tool-agentic data from real-world mcp environments},
  author={Xu, Zhangchen and Soria, Adriana Meza and Tan, Shawn and Roy, Anurag and Agrawal, Ashish Sunil and Poovendran, Radha and Panda, Rameswar},
  journal={arXiv preprint arXiv:2510.01179},
  year={2025}
}

@article{agentic_rl_survey,
  title={The landscape of agentic reinforcement learning for llms: A survey},
  author={Zhang, Guibin and Geng, Hejia and Yu, Xiaohang and Yin, Zhenfei and Zhang, Zaibin and Tan, Zelin and Zhou, Heng and Li, Zhongzhi and Xue, Xiangyuan and Li, Yijiang and others},
  journal={arXiv preprint arXiv:2509.02547},
  year={2025}
}

@article{qu2025tool,
  title={Tool learning with large language models: A survey},
  author={Qu, Changle and Dai, Sunhao and Wei, Xiaochi and Cai, Hengyi and Wang, Shuaiqiang and Yin, Dawei and Xu, Jun and Wen, Ji-Rong},
  journal={Frontiers of Computer Science},
  volume={19},
  number={8},
  pages={198343},
  year={2025},
  publisher={Springer}
}

@article{demystify_agentic_rl,
  title={Demystifying reinforcement learning in agentic reasoning},
  author={Yu, Zhaochen and Yang, Ling and Zou, Jiaru and Yan, Shuicheng and Wang, Mengdi},
  journal={arXiv preprint arXiv:2510.11701},
  year={2025}
}

@article{roll,
  title={Reinforcement Learning Optimization for Large-Scale Learning: An Efficient and User-Friendly Scaling Library},
  author={Wang, Weixun and Xiong, Shaopan and Chen, Gengru and Gao, Wei and Guo, Sheng and He, Yancheng and Huang, Ju and Liu, Jiaheng and Li, Zhendong and Li, Xiaoyang and others},
  journal={arXiv preprint arXiv:2506.06122},
  year={2025}
}

@article{qwen3,
  title={Qwen3 technical report},
  author={Yang, An and Li, Anfeng and Yang, Baosong and Zhang, Beichen and Hui, Binyuan and Zheng, Bo and Yu, Bowen and Gao, Chang and Huang, Chengen and Lv, Chenxu and others},
  journal={arXiv preprint arXiv:2505.09388},
  year={2025}
}

@inproceedings{Cheng2025TravelBenchAR,
  title={TravelBench: A Real-World Benchmark for Multi-Turn and Tool-Augmented Travel Planning},
  author={Xiang Cheng and Yulan Hu and Xiangwen Zhang and Lu Xu and Zheng Pan and Xin Li and Yong Liu},
  year={2025},
  url={https://api.semanticscholar.org/CorpusID:284313793}
}

@inproceedings{gunjal2025rubrics,
  title={Rubrics as Rewards: Reinforcement Learning Beyond Verifiable Domains},
  author={Gunjal, Anisha and Wang, Anthony and Lau, Elaine and Nath, Vaskar and He, Yunzhong and Liu, Bing and Hendryx, Sean M},
  booktitle={NeurIPS 2025 Workshop on Efficient Reasoning},
  year={2025},
}

@inproceedings{hashemi2024llm,
  title={LLM-Rubric: A Multidimensional, Calibrated Approach to Automated Evaluation of Natural Language Texts},
  author={Hashemi, Helia and Eisner, Jason and Rosset, Corby and Van Durme, Benjamin and Kedzie, Chris},
  booktitle={Proceedings of the 62nd Annual Meeting of the Association for Computational Linguistics (Volume 1: Long Papers)},
  pages={13806--13834},
  year={2024}
}

@misc{TIR_understand,
      title={Understanding Tool-Integrated Reasoning}, 
      author={Heng Lin and Zhongwen Xu},
      year={2025},
      eprint={2508.19201},
      archivePrefix={arXiv},
      primaryClass={cs.LG},
      url={https://arxiv.org/abs/2508.19201}, 
}

@misc{stma,
      title={STMA: A Spatio-Temporal Memory Agent for Long-Horizon Embodied Task Planning}, 
      author={Mingcong Lei and Yiming Zhao and Ge Wang and Zhixin Mai and Shuguang Cui and Yatong Han and Jinke Ren},
      year={2025},
      eprint={2502.10177},
      archivePrefix={arXiv},
      primaryClass={cs.AI},
      url={https://arxiv.org/abs/2502.10177}, 
}

@article{agentic_limits,
  title={What Limits Agentic Systems Efficiency?},
  author={Bian, Song and Yan, Minghao and Jayarajan, Anand and Pekhimenko, Gennady and Venkataraman, Shivaram},
  journal={arXiv preprint arXiv:2510.16276},
  year={2025}
}

@article{difficulty_know,
  title={Know when to explore: Difficulty-aware certainty as a guide for llm reinforcement learning},
  author={Li, Ang and Yuan, Zhihang and Zhang, Yang and Liu, Shouda and Wang, Yisen},
  journal={arXiv preprint arXiv:2509.00125},
  year={2025}
}

@article{weak2strong,
  title={Weak-to-strong generalization: Eliciting strong capabilities with weak supervision},
  author={Burns, Collin and Izmailov, Pavel and Kirchner, Jan Hendrik and Baker, Bowen and Gao, Leo and Aschenbrenner, Leopold and Chen, Yining and Ecoffet, Adrien and Joglekar, Manas and Leike, Jan and others},
  journal={arXiv preprint arXiv:2312.09390},
  year={2023}
}

@inproceedings{acebench,
    title = "{ACEB}ench: A Comprehensive Evaluation of {LLM} Tool Usage",
    author = "Chen, Chen  and
      Hao, Xinlong  and
      Liu, Weiwen  and
      Huang, Xu  and
      Zeng, Xingshan  and
      Yu, Shuai  and
      Li, Dexun  and
      Huang, Yuefeng  and
      Liu, Xiangcheng  and
      Xinzhi, Wang  and
      Liu, Wu",
    editor = "Christodoulopoulos, Christos  and
      Chakraborty, Tanmoy  and
      Rose, Carolyn  and
      Peng, Violet",
    booktitle = "Findings of the Association for Computational Linguistics: EMNLP 2025",
    month = nov,
    year = "2025",
    address = "Suzhou, China",
    publisher = "Association for Computational Linguistics",
    url = "https://aclanthology.org/2025.findings-emnlp.697/",
    doi = "10.18653/v1/2025.findings-emnlp.697",
    pages = "12970--12998",
    ISBN = "979-8-89176-335-7"
}

@misc{tau2bench,
      title={$\tau^2$-Bench: Evaluating Conversational Agents in a Dual-Control Environment}, 
      author={Victor Barres and Honghua Dong and Soham Ray and Xujie Si and Karthik Narasimhan},
      year={2025},
      eprint={2506.07982},
      archivePrefix={arXiv},
      primaryClass={cs.AI},
      url={https://arxiv.org/abs/2506.07982}, 
}

@inproceedings{bfcl,
    title={The Berkeley Function Calling Leaderboard ({BFCL}): From Tool Use to Agentic Evaluation of Large Language Models},
    author={Shishir G Patil and Huanzhi Mao and Fanjia Yan and Charlie Cheng-Jie Ji and Vishnu Suresh and Ion Stoica and Joseph E. Gonzalez},
    booktitle={Forty-second International Conference on Machine Learning},
    year={2025},
    url={https://openreview.net/forum?id=2GmDdhBdDk}
}

@misc{aime,
  author       = {{AIME}},
  title        = {{AIME problems and solutions}},
  year         = {2025},
  howpublished = {\url{https://artofproblemsolving.com/wiki/index.php/AIME_Problems_and_Solutions}}
}

@misc{livecodebench,
      title={LiveCodeBench: Holistic and Contamination Free Evaluation of Large Language Models for Code}, 
      author={Naman Jain and King Han and Alex Gu and Wen-Ding Li and Fanjia Yan and Tianjun Zhang and Sida Wang and Armando Solar-Lezama and Koushik Sen and Ion Stoica},
      year={2024},
      eprint={2403.07974},
      archivePrefix={arXiv},
      primaryClass={cs.SE},
      url={https://arxiv.org/abs/2403.07974}, 
}

@article{arenahard,
  title={From Crowdsourced Data to High-Quality Benchmarks: Arena-Hard and BenchBuilder Pipeline},
  author={Li, Tianle and Chiang, Wei-Lin and Frick, Evan and Dunlap, Lisa and Wu, Tianhao and Zhu, Banghua and Gonzalez, Joseph E and Stoica, Ion},
  journal={arXiv preprint arXiv:2406.11939},
  year={2024}
}

@article{ifeval,
  title={Instruction-Following Evaluation for Large Language Models},
  author={Zhou, Jeffrey and Lu, Tianjian and Mishra, Swaroop and Brahma, Siddhartha and Basu, Sujoy and Luan, Yi and Zhou, Denny and Hou, Le},
  journal={arXiv preprint arXiv:2311.07911},
  year={2023}
}

@misc{mmlupro,
      title={MMLU-Pro: A More Robust and Challenging Multi-Task Language Understanding Benchmark}, 
      author={Yubo Wang and Xueguang Ma and Ge Zhang and Yuansheng Ni and Abhranil Chandra and Shiguang Guo and Weiming Ren and Aaran Arulraj and Xuan He and Ziyan Jiang and Tianle Li and Max Ku and Kai Wang and Alex Zhuang and Rongqi Fan and Xiang Yue and Wenhu Chen},
      year={2024},
      eprint={2406.01574},
      archivePrefix={arXiv},
      primaryClass={cs.CL},
      url={https://arxiv.org/abs/2406.01574}, 
}

@misc{ceval,
      title={C-Eval: A Multi-Level Multi-Discipline Chinese Evaluation Suite for Foundation Models}, 
      author={Yuzhen Huang and Yuzhuo Bai and Zhihao Zhu and Junlei Zhang and Jinghan Zhang and Tangjun Su and Junteng Liu and Chuancheng Lv and Yikai Zhang and Jiayi Lei and Yao Fu and Maosong Sun and Junxian He},
      year={2023},
      eprint={2305.08322},
      archivePrefix={arXiv},
      primaryClass={cs.CL},
      url={https://arxiv.org/abs/2305.08322}, 
}
\bibliographystyle{iclr2026_conference}

\appendix
\section{Tool Schemas and Prompt Templates} \label{app:prompt}
\subsection{Tool schemas}~\label{app:tools}

\paragraph{1. Map \& Navigation Tools}
\begin{itemize}
    \item \textbf{map\_search\_places}: Searches for Points of Interest (POIs) based on keywords, categories, or addresses. Supports nearby search with customizable radius, administrative region filtering, and multiple sorting options (distance, rating, price). Returns rich metadata including ratings, prices, business hours, and user reviews.
    \item \textbf{map\_compute\_routes}: Computes optimal routes from origin to destination with optional waypoints. Supports six transport modes: driving, walking, cycling, public transit, motorcycle, and truck. Provides optional real-time traffic awareness and route preferences (e.g., avoid tolls, prefer highways).
    \item \textbf{map\_search\_along\_route}: Finds POIs within a buffer zone along a planned route, ideal for scenarios like ``find a gas station on my way home.'' Automatically plans the base route and searches within the specified corridor.
    \item \textbf{map\_search\_central\_places}: Locates places that are convenient for multiple parties by computing distance metrics to all origins. Supports three optimization strategies: balanced (overall optimal), minimize maximum distance (fairness-oriented), and minimize total distance (efficiency-oriented).
    \item \textbf{map\_search\_ranking\_list}: Retrieves curated ranking lists such as ``Must-Eat List'' or ``Black Pearl'' for a specified category and region. Returns ranked POIs with detailed recommendation reasons and featured tags.
\end{itemize}

\paragraph{2. Travel \& Transportation Tools}
\begin{itemize}
    \item \textbf{travel\_search\_flights}: Searches domestic flight information between cities. Supports optional multi-day queries for price comparison across consecutive dates. Returns flight numbers, airlines, departure/arrival times, aircraft types, and price ranges.
    \item \textbf{travel\_search\_trains}: Queries train and high-speed rail schedules between cities. Supports multi-day queries for schedule comparison. Returns train numbers, schedules, duration, stations, and ticket prices.
\end{itemize}

\paragraph{3. Weather Tools}
\begin{itemize}
    \item \textbf{weather\_current\_conditions}: Retrieves real-time weather conditions for a specified location, including temperature, feels-like temperature, weather phenomena, wind direction/speed, and Air Quality Index (AQI).
    \item \textbf{weather\_forecast\_days}: Provides multi-day weather forecasts (up to 5 additional days) for a given location. Supports both specific date queries and date range queries.
\end{itemize}

\paragraph{4. Information Retrieval Tools}
\begin{itemize}
    \item \textbf{web\_search}: Performs open-domain web search for general knowledge, real-time news, historical events, and policy information that falls outside the spatio-tempora domain.
\end{itemize}

\subsection{User Instruction Annotation Prompt Template}
\label{app:intent_classification_prompt}

\begin{lstlisting}[language=XML, basicstyle=\ttfamily\scriptsize,showstringspaces=false, breaklines=true, frame=single, caption={Prompt template used for user instruction classification.}]
# Amap User Instruction Annotation System

## Role And Task

You are an expert annotation specialist for Amap user instructions. Your task is to analyze user instructions submitted to Amap (text or voice transcribed via ASR) and classify them using a comprehensive, hierarchical taxonomy system.

**Your Objectives:**
1. Identify the PRIMARY INTENT that represents the user's main goal
2. Identify up to 3 SECONDARY INTENTS that represent additional goals (if applicable)
3. Annotate ALL EXPLICIT CONSTRAINTS from the 5 auxiliary dimensions
4. For Route Planning and Itinerary Planning intents ONLY: specify departure/arrival time constraints in temporal_details

**Key Principles:**
- Base all annotations on the explicit content of the instruction
- Use geospatial and temporal common sense (e.g., approximate adjacency between cities, typical trip durations) to infer real user intent
- Do not invent or assume new constraints (budget, preferences, time limits, etc.) that are not explicitly stated
- Focus on the user's core action/goal when selecting primary intent
- Use fully-qualified hierarchical IDs of the leaf intent for both primary and auxiliary dimensions

---

## Input and Output Format

### Input Format

Each task provides a single user instruction wrapped in `instruction` tags.

```
<instruction>
{instruction}
</instruction>
```

### Output Format

Your output MUST follow this XML structure exactly and contain ONLY this XML block:

```xml
<primary_intent>
intent_id_with_full_path
</primary_intent>

<secondary_intents>
secondary_intent_id_1
secondary_intent_id_2
secondary_intent_id_3
</secondary_intents>

<auxiliary_dimensions>
dimension_id_1
dimension_id_2
dimension_id_3
</auxiliary_dimensions>

<temporal_details>
<departure>
temporal_constraint_id
</departure>
<arrival>
temporal_constraint_id
</arrival>
</temporal_details>
```

---
...

[Table of User Intent Taxonomy]

...
---

## Annotation Guidelines

### Multi-Intent Annotation Framework

**Structure:**
- **Primary Intent** (Required): The main goal driving the instruction. Choose the best-matched leaf intent from the intents defined in "Primary Dimension: User Intent Taxonomy" section.
- **Secondary Intents** (Optional): Up to 3 supporting leaf intents that complement the primary intent. List in order of importance.
- **Auxiliary Dimensions** (Optional): ALL explicit leaf constraints mentioned in the instruction.
- **Temporal Details** (Optional): Only output the `<temporal_details>` block when the instruction involves departure/arrival timing or trip start/end concepts. This is especially important for `planning_and_decision.route_planning` and `planning_and_decision.itinerary_planning` intents, which typically involve such temporal information. If no departure/arrival timing is mentioned, omit the entire block.

### Systematic Annotation Process

Follow this systematic approach for every annotation:

**Step 1: Identify Core Action/Goal**
- What is the primary action the user wants to perform?
- Search/Discover? Plan/Route? Compare? Query info? Navigate? Check rules?
- Focus on explicit action verbs and functional goals, NOT contextual hints or scenario keywords
- Ask: "What would the system need to do to fulfill this instruction?"

**Step 2: Select Primary Intent**
- Based on the core action, select the specific leaf intent from "Primary Dimension: User Intent Taxonomy" section
- **Priority Rule:** 
  - (1) More specific > More general
  - (2) Action-oriented > Information-oriented 
  - (3) Best supported by explicit constraints
- Always specify the most appropriate leaf intent
- **Out-of-domain:** If clearly non-travel/incomprehensible/pure greetings goes to `other_or_unclear` (and keep all lists empty)

**Step 3: Check Exclusion Features**
- Verify the instruction does NOT match any "NOT" exclusion criteria in taxonomy definitions
- Use "Critical Distinctions and Edge Cases" section to resolve common confusions
- Ensure contextual hints have not misled the classification

**Step 4: Identify Secondary Intents**
- Determine if the instruction expresses additional goals beyond the primary intent
- Annotate up to 3 secondary intents ordered by importance
- Only annotate what is clearly expressed or strongly implied
- Order by importance relative to the primary goal

**Step 5: Annotate Auxiliary Dimensions**
- Extract ALL explicit constraints from the instruction
- Use fully-qualified IDs (e.g., `spatial_constraints.within_city.nearby`)
- Multiple values within same dimension are allowed
- Use basic geospatial/temporal common sense to map explicit mentions to taxonomy buckets

**Step 6: Add Temporal Details (When Applicable)**
- Only output `<temporal_details>` block if instruction involves departure/arrival timing or trip start/end concepts
- Pay special attention for Route Planning and Itinerary Planning intents (typically involve temporal information)
- Use IDs from `temporal_constraints` hierarchy
- **Departure:** When user wants to start the journey or start the trip
- **Arrival:** When user needs to reach the destination or finish the trip
- **If either departure or arrival has no value**: Omit that field entirely (do not output empty tags)
- Use `temporal_constraints.fuzzy_time.flexible` only when user explicitly states no time constraints
- **If no temporal information**: Omit the entire `<temporal_details>` block

---

## Annotation Examples and Critical Distinctions

This section provides complete annotation examples and highlights critical distinctions between commonly confused intents.
...

---

### Critical Distinctions and Edge Cases

This section highlights the most commonly confused intent pairs and critical decision points.
...
\end{lstlisting}

\subsection{Difficulty Scoring Annotation Prompt Template}
\label{app:difficulty_scoring_prompt}
\begin{lstlisting}[language=XML, basicstyle=\ttfamily\scriptsize,showstringspaces=false, breaklines=true, frame=single, caption={Prompt template used for difficulty scoring.}]
# Role
You are an expert **Amap Agent Simulator and Difficulty Evaluator**. Your objective is to assess the difficulty of a user's query by first **simulating** the execution plan using a strict set of tools, and then **scoring** the complexity based on that simulation.

# Input Data
You will receive:
1.  **User Query**: The user's natural language command.
2.  **Context**: Current location, time, user profile, etc.
3.  **Tool Definitions**: A list of available tools will be given in Appendix.

# Task Workflow

## Step 1: Intent & Context Analysis (Mental Sandbox)
Analyze the user's intent and available context. Determine if the task is feasible.
* **Feasibility Check**: Do you have the *necessary and sufficient conditions* to start planning?
* **Missing Information**: Distinguish between "Contextual Missing" (e.g., "Go *there*" with no reference -> Unexecutable) vs "Retrievable Missing" (e.g., "Find *nearest gas station*" -> Executable via search).
* **Tool Coverage**: Can the request be fulfilled using ONLY the provided tools?

## Step 2: Simulation Planning (The "Dry Run")
If the task is feasible, generate a logical **Tool Chain**.
* **Dependency**: Ensure Step B can only happen after Step A if B needs A's output.
* **Data Flow**: Explicitly state where parameters come from (e.g., `$Context.location`, `$Step1.poi_id`).
* **Logic**: Describe branching logic for complex scenarios (e.g., "If tickets available, then Book; else Waitlist").
* **Anti-Hallucination**:
    * **Strict Toolset**: Use ONLY tools defined in `<tools>`. If a needed tool doesn't exist, mark as `MissingTool`.
    * **No Magic Data**: Do not invent coordinates, POI IDs, or user preferences. If they aren't in Context or previous tool outputs, you must plan a tool call to get them.
    * **Time Awareness**: Use the provided `User Query Time` as the anchor for all temporal queries.

## Step 3: Difficulty Scoring (Based on Simulation)
Assign a score from **-1 to 5** based on the simulation experience.

# Scoring Rubric

## Special Categories (Score: -1 or 0)

### Score -1: Unexecutable
The query cannot be solved by the agent regardless of reasoning capabilities.
* **Case A (MissingInfo):** Critical entities are missing, and context/history cannot resolve them. (e.g., "Go to that mall" with no context).
* **Case B (MissingTool):** The intent is clear, but the functionality is outside the agent's scope. (e.g., "Change the backend code of Google Maps", "Play a video").
* **Note:** If a query falls here, stop analysis and assign -1.

### Score 0: No Tool Required
The query is chitchat, general knowledge, or a greeting that requires no geospatial tool execution. (e.g., "Hello", "Who are you?", "Tell me a joke").

---

## Graded Difficulty (Score: 1 to 5)
For executable queries, assign a score based on the **Orthogonal Matrix** of Cognitive Load (Tool Selection) and Mechanical Load (Execution Steps).

### Score 1: Very Easy (Atomic)
* **Tool Selection:** **Trivial**. The query explicitly keywords a specific tool or intent (e.g., "Navigate to...", "Search for..."). No reasoning required.
* **Execution:** Single-step execution (1 turn). Uses 1 tool type.
* **Query Quality:** Clear and unambiguous.
* **Example:** "Navigate to the Eiffel Tower." / "What is the weather in London?"

### Score 2: Easy (Linear Chain)
* **Tool Selection:** **Straightforward**. Requires mapping a standard intent to a standard tool sequence. The path is linear.
* **Execution:** Short sequence (2 turns). Uses 1-2 tool types. Typically involves `Search` -> `Action` (e.g., Navigate).
* **Query Quality:** Good.
* **Example:** "Find a gas station nearby and take me there." (Search -> Select -> Navigate)

### Score 3: Medium (Conditional / Parameterized)
* **Tool Selection:** **Moderate**. Requires analyzing constraints or filters. The agent must extract specific parameters (price, rating, open status) to configure the tools correctly.
* **Execution:** Medium sequence (3 turns). Uses 1-3 tool types.  Involves filtering, sorting, or "Search Along Route" logic.
* **Query Quality:** Average to Good. May require slight inference.
* **Example:** "Find a cheap Italian restaurant on the way to the airport that is open now." (Route planning -> Search along route -> Filter by price/cuisine/time).

### Score 4: Hard (Multi-Intent / Optimization)
* **Tool Selection:** **Challenging**. The query contains multiple distinct sub-goals or requires comparison/optimization logic. The agent must decompose the query into parallel or complex serial tasks.
* **Execution:** Long sequence (4-6 turns). Uses multiple tool types (3+ types).
* **Query Quality:** May contain implicit requirements or complex sentence structures.
* **Example:** "Plan a date night: first a movie at a cinema with good ratings, then a nearby bar, and finally drive me home. Avoid highways." (Multi-stage planning + Preferences).

### Score 5: Very Hard (Complex Reasoning / Edge Cases)
* **Tool Selection:** **Expert Level**. The user's intent is abstract, highly implicit, or requires cross-referencing multiple domains. The agent must "invent" a solution path using the tools creatively.
* **Execution:** Massive sequence (7+ turns). High tool variety.
* **Query Quality:** Poor/Ambiguous (requiring deep inference) OR Excellent but extremely complex constraints.
* **Example:** "I have 3 hours to kill before my flight at JFK. Find me a scenic spot to read a book within 20 mins drive, get me coffee on the way, and make sure I don't hit traffic coming back." (Time budgeting + Traffic prediction + Multi-stop + Vague "scenic" definition).

---

# Output Format

Your output MUST follow the structure below exactly. It consists of an `<analysis>` XML block followed by a `<response>` block.

```xml
<analysis>
    <intent_analysis>
        <intent>Brief description of user intent</intent>
        <feasibility>Executable | MissingInfo | MissingTool | NoToolNeeded</feasibility>
        <missing_details>Describe what is missing (if any)</missing_details>
    </intent_analysis>

    <simulation>
        <step id="1">
            <tool_name>tool_name_here</tool_name>
            <reason>Why this tool is needed</reason>
            <parameters>
                <param name="arg_name">Source (e.g., $Context.lat or 'gas station')</param>
            </parameters>
        </step>
        <step id="2">
            <tool_name>tool_name_here</tool_name>
            <parameters>
                <param name="arg_name">$Step1.result.id</param>
            </parameters>
        </step>
    </simulation>

    <scoring>
        <tool_selection_difficulty>
            <rating>very easy | easy | medium | hard | very hard</rating>
            <reasoning>
                Explain the cognitive load required to select these tools. Was it obvious? Did it require inferring implicit constraints?
            </reasoning>
        </tool_selection_difficulty>
        
        <execution_complexity>
            <estimated_turns>Integer (e.g., 3)</estimated_turns>
            <tool_variety>Integer (e.g., 2 types)</tool_variety>
        </execution_complexity>

        <query_quality>
            <rating>poor | average | good | excellent</rating>
            <impact>How quality affected the difficulty</impact>
        </query_quality>

        <final_score>Integer (-1 to 5)</final_score>
    </scoring>
</analysis>
<response>
    * MissingInfo: [[true/false]]
    * MissingTool: [[true/false]]
    Rating: [[X]]
</response>
```
-----
## Appendix: List of Available Tools
You have access to the following tools definitions to assist with your simulation.

<tools>
{tools_json_schema}
</tools>

\end{lstlisting}

\subsection{Reward Model Prompt Template}
\label{app:reward_prompt}

The following prompt illustrates the dynamic evaluation logic used to score \amapagent trajectories. The evaluator first classifies the task type to assign weights, checks strictly for hallucinations, and then provides a reasoned score.

\begin{lstlisting}[language=XML, basicstyle=\ttfamily\scriptsize, breaklines=true, frame=single, caption={Prompt template with dynamic weighting and hallucination veto logic.}]
## Role
You are a senior AI Agent Evaluation Expert. Evaluate the interaction based on tool usage, reasoning, and the final answer.

## Evaluation Criteria (Summary)
1. Reasoning & Proactivity: 
   - Does the model correct user errors proactively (e.g., fixing location mismatches) instead of rejecting them?
   - Are tool arguments precise without redundancy?
2. Information Fidelity:
   - STRICT VETO: Any hallucinated fact (price, time, distance) results in a Final Score of 0.
   - Must strictly ground answers in <tool_response>.
3. Presentation:
   - Does it close the service loop with actionable next steps?
   - Is the format structured and user-friendly?

## Dynamic Scoring Logic
Step 1: Determine Weights (w_reasoning + w_integration + w_presentation = 1.0)
   - Scenario A (Complex Planning): Focus on Reasoning (e.g., 0.6 / 0.3 / 0.1)
   - Scenario B (Data Retrieval): Focus on Fidelity (e.g., 0.2 / 0.6 / 0.2)
   - Scenario C (Consultation): Focus on Presentation (e.g., 0.3 / 0.3 / 0.4)

Step 2: Calculate Score
   - final_score = (Rating1 * w1) + (Rating2 * w2) + (Rating3 * w3)
   - If <has_hallucination> is true, final_score = 0.0

## Output Format
Strictly output raw XML without Markdown blocks.

```xml
<evaluation_report>
    <hallucination_analysis>
        <has_hallucination></has_hallucination>
        <details></details>
    </hallucination_analysis>
    <weight_analysis>
        <rationale></rationale>
        <weights>
            <w_reasoning>...</w_reasoning>
            <w_integration>...</w_integration>
            <w_presentation>...</w_presentation>
        </weights>
    </weight_analysis>
    <dimension_reasoning>
        <rationale>...</rationale>
        <rating></rating>
    </dimension_reasoning>
    <dimension_integration>
        <rationale>...</rationale>
        <rating>...</rating>
    </dimension_integration>
    <dimension_presentation>
        <rationale>...</rationale>
        <rating>...</rating>
    </dimension_presentation>
    <final_score></final_score>
</evaluation_report>
\end{lstlisting}

\section{Case Studies} \label{app:casestudy}

% ==========================================
% 在这里自定义颜色
% ==========================================

% 方式 1: 使用 RGB 数值 (0-255) - 比如淡黄色
% \definecolor{mycolor}{RGB}{255, 240, 200} 

% 方式 2: 使用 HTML 十六进制代码 - 比如淡蓝色 (类似 CSS)
\definecolor{mycolor}{HTML}{eff6fd}
\definecolor{color_amapblue}{HTML}{3472ed}
\definecolor{color_amaplightblue}{HTML}{87e1f9}
\definecolor{color_amapgreen}{HTML}{c1fca0}
\definecolor{color_amapgrey}{HTML}{e5eff0}

In this section, we conduct a case study to demonstrate our \model...
% =================================================================
% 修改说明：
% 1. colframe=color_amaplightblue : 将边框和标题背景改为你的淡蓝色
% 2. coltitle=black : (可选) 因为淡蓝色背景比较亮，建议将标题文字设为黑色，否则默认的白色看不清
% =================================================================

\begin{CJK*}{UTF8}{gkai} % 开始中文环境 (楷体)

% ==================================================================================
% Example 1 (您之前的 Case，增加了 Response 渲染效果)
% ==================================================================================

\subsection{Example 1: Multi-step Geospatial Reasoning (Route-based Search)}
\begin{tcolorbox}[
    breakable,
    enhanced,
    colframe=color_amapblue,
    colback=mycolor,
    coltitle=white,
    title = {\textbf{Case Study: Route-based Dining Recommendation}},
    fonttitle=\bfseries
]

\textbf{Query:} 我现在在海口的云洞图书馆，我住在楠丰酒店，请问我们回酒店的路上有什么好吃的\\
\textbf{Intent:} discovery.poi\_search.spatial\_optimized\_poi\_search
\tcblower

% --- Step 1 ---
\textbf{Action Step 1: Entity Resolution}
\begin{itemize}[leftmargin=12pt, itemsep=0pt]
  \item \textbf{Tool:} \texttt{map\_search\_places} (called twice in parallel)
  \item \textbf{Arguments:} 
    \begin{itemize}[leftmargin=10pt]
        \item Query A: \{``query'': ``云洞图书馆'', ``region'': ``海口市''\}
        \item Query B: \{``query'': ``楠丰酒店'', ``region'': ``海口市''\}
    \end{itemize}
  \item \textbf{Observation:} 
    \begin{itemize}[leftmargin=10pt]
        \item Origin: Cloud Cave Library ($20.046661, 110.316404$)
        \item Destination: Nanfeng Hotel ($20.026726, 110.232278$)
    \end{itemize}
\end{itemize}
\vspace{5pt}
% --- Step 2 ---
\textbf{Action Step 2: Search Along Route}
\begin{itemize}[leftmargin=12pt, itemsep=0pt]
  \item \textbf{Tool:} \texttt{map\_search\_along\_route}
  \item \textbf{Arguments:} 
    \{``origin'': $P_{origin}$, ``destination'': $P_{dest.}$, ``query'': ``美食'', ``open\_now'': true\}
  \item \textbf{Observation:} Found 49 places. Top results:
    \begin{enumerate}[leftmargin=15pt, label=\arabic*.]
        \item Gu Gu Gu Hainan Cuisine (Scenic View)
        \item Tanmen Fisherman Seafood Hotpot
        \item Kejia Ling Fusion Cuisine
    \end{enumerate}
\end{itemize}

\vspace{5pt}

% --- Response Generator (模拟最终渲染效果) ---
\textbf{Response Generator:}
\begin{tcolorbox}[colback=white, colframe=color_amapblue!30!white, arc=4pt, boxrule=1pt, left=4pt, right=4pt, top=4pt, bottom=4pt]
    \small
    % BLUF 区域
    \textbf{BLUF:} 回酒店路上有多个美食选择，推荐以下热门餐厅，均在路线附近且营业中。
    
    \par\noindent\rule{\textwidth}{0.4pt} % 分割线
    
    % 核心推荐列表 (使用 faUtensils 图标)
    \textbf{\textcolor{color_amapblue}{\faUtensils\ 推荐美食地点}}
    \begin{enumerate}[leftmargin=15pt, itemsep=4pt, topsep=2pt]
    \item \textbf{古姑姑·品质海南菜私房餐厅}
    \begin{itemize}[leftmargin=10pt, label={\tiny\textbullet}, itemsep=0pt, parsep=0pt]
        \item \textbf{位置：}秀英区滨海大道假日海滩游客中心二楼
        \item \textbf{营业时间：}10:00-13:30, 16:30-21:30
        \item \textbf{特色：}270°观海视野，海南菜经典，秀英区观景餐厅榜第3名
        \item \textbf{提示：}适合晚餐，海景+美食双重享受。
    \end{itemize}

    \item \textbf{潭门渔夫海鲜火锅}
    \begin{itemize}[leftmargin=10pt, label={\tiny\textbullet}, itemsep=0pt, parsep=0pt]
        \item \textbf{位置：}龙华区上邦百汇城1号楼3楼
        \item \textbf{营业时间：}11:30-23:00
        \item \textbf{特色：}海鲜火锅，海口市海鲜火锅榜第2名
        \item \textbf{提示：}适合多人聚餐，食材新鲜。
    \end{itemize}

    \item \textbf{客家岭湘粤融合菜}
    \begin{itemize}[leftmargin=10pt, label={\tiny\textbullet}, itemsep=0pt, parsep=0pt]
        \item \textbf{位置：}秀英区滨海大道80号幸福里小区
        \item \textbf{营业时间：}10:00-14:00, 17:00-21:30
        \item \textbf{特色：}客家+粤菜融合，丽晶路美食榜第3名
        \item \textbf{提示：}适合偏好清淡口味的旅客。
    \end{itemize}
\end{enumerate}
    % \begin{enumerate}[leftmargin=12pt, itemsep=2pt, topsep=2pt]
    %     % 餐厅 1
    %     \item \textbf{古姑姑·品质海南菜} (秀英区)
    %     \begin{itemize}[leftmargin=6pt, label={}, itemsep=0pt, parsep=0pt]
    %         \item \scriptsize \textcolor{gray}{10:00-21:30 | 观景餐厅 | 秀英区榜第3名}
    %     \end{itemize}
        
    %     % 餐厅 2
    %     \item \textbf{潭门渔夫海鲜火锅} (龙华区)
    %     \begin{itemize}[leftmargin=6pt, label={}, itemsep=0pt, parsep=0pt]
    %         \item \scriptsize \textcolor{gray}{11:30-23:00 | 鲜香浓郁 | 海口火锅榜第2名}
    %     \end{itemize}
        
    %     % 餐厅 3 & 4
    %     \item \textbf{其他热门：}
    %     \begin{itemize}[leftmargin=6pt, label={}, itemsep=0pt, parsep=0pt]
    %         \item \scriptsize \textbf{太二酸菜鱼：}知名品牌，酸辣开胃 (至21:00)
    %         \item \textbf{梁庄琼味酒楼：}滨贸路榜第1名，本地推荐
    %     \end{itemize}
    % \end{enumerate}
    
    \par\noindent\rule{\textwidth}{0.4pt}
    
    % 下一步建议 (使用 faMapMarker/faThumbtack 图标)
    \textbf{\textcolor{color_amapblue}{\faMapMarker*\ 下一步建议}}
    \begin{itemize}[leftmargin=12pt, itemsep=0pt, topsep=2pt]
        \item \textbf{导航：}使用高德地图导航至“古姑姑”或“潭门渔夫”。
        \item \textbf{提示：}若时间较晚，建议选择营业至23:00的潭门渔夫。
    \end{itemize}
    
    \vspace{3pt}
    
    % 底部 Tip (使用 faCar 图标)
    \colorbox{color_amapgrey}{\parbox{\dimexpr\linewidth-6pt}{
        \scriptsize 
        \textbf{\faCar\ 路线提示：}从云洞图书馆出发沿滨海大道行驶，上述餐厅均在路线两侧 \textbf{2公里内} (驾车约10-15分钟)。
    }}
\end{tcolorbox}

\end{tcolorbox}

\subsection{Example 2: Complex Multi-Point Itinerary Assessment}
\begin{tcolorbox}[
    breakable,
    enhanced,
    colframe=color_amapblue,
    colback=mycolor,
    coltitle=white,
    title = {\textbf{Case Study: Multi-stop Route Feasibility Check}},
    fonttitle=\bfseries
]

% --- 用户 Query ---
\textbf{Query:} 广州海珠广场去广东博物馆，陈家祠，珠江夜游，同福路，永庆坊，沙面岛路线方便么\\
\textbf{Intent:} planning\_and\_decision.itinerary\_planning.simple\_route\_itinerary
\tcblower

% --- Step 1: 批量实体解析 ---
\textbf{Action Step 1: Batch Entity Resolution}
\begin{itemize}[leftmargin=12pt, itemsep=0pt]
  \item \textbf{Tool:} \texttt{map\_search\_places} (called 7 times in parallel)
  \item \textbf{Observation:} Successfully resolved coordinates for all locations:
    \begin{itemize}[leftmargin=10pt, itemsep=0pt]
        \item Origin: Haizhu Square ($23.114, 113.266$)
        \item Intermediates: Chen Clan Ancestral Hall (Liwan), Tianzi Wharf (Yuexiu), etc.
        \item Destination: Guangdong Museum ($23.106,113.244$)
    \end{itemize}
\end{itemize}
\vspace{5pt}
% --- Step 2: 路线计算 (基于真实数据) ---
\textbf{Action Step 2: Route Computation}
\begin{itemize}[leftmargin=12pt, itemsep=0pt]
  \item \textbf{Tool:} \texttt{map\_compute\_routes}
  \item \textbf{Arguments:} \{``origin'': $P_{Haizhu}$, ``destination'': $P_{Museum}$, ``intermediates'': [$P_{Chen}, P_{Tianzi}...$], ``modes'': [``driving'']\}
  \item \textbf{Observation:} 
    \begin{itemize}[leftmargin=10pt]
        \item \textbf{Metrics:} 38.3 km | 2h 28m | \textbf{84 Traffic Lights}.
        \item \textbf{Congestion:} 7.1 km congestion, specifically at \textit{Dongfeng West Rd} (+6 min delay).
    \end{itemize}
\end{itemize}

\vspace{5pt}

% --- Response Generator (模拟最终渲染效果) ---
\textbf{Response Generator (Rendered Output):}
\begin{tcolorbox}[colback=white, colframe=color_amapblue!30!white, arc=4pt, boxrule=1pt, left=4pt, right=4pt, top=4pt, bottom=4pt]
    \small
    % BLUF
    \textbf{BLUF:} 路线可行但效率较低（需2.5小时，84个红绿灯）。建议避开晚高峰或改用地铁。
    
    \par\noindent\rule{\textwidth}{0.4pt}
    
    % 路况分析 (使用 faTrafficLight 图标)
    \textbf{\textcolor{color_amapblue}{\faTrafficLight\ 路况核心数据}}
    \begin{itemize}[leftmargin=12pt, itemsep=2pt, topsep=2pt]
        \item \textbf{耗时预警：} 全程约 2小时28分钟，途经 **84个红绿灯**。
        \item \textbf{拥堵黑点：} **东风西路** 拥堵严重，预计延误 6分钟以上。
        \item \textbf{路线特点：} 跨度大（天河$\leftrightarrow$荔湾），晚高峰(17:00-19:00)通行困难。
    \end{itemize}
    
    \par\noindent\rule{\textwidth}{0.4pt}
    
    % 优化建议 (使用 faRoute 图标)
    \textbf{\textcolor{color_amapblue}{\faRoute\ 行程优化建议}}
    \begin{enumerate}[leftmargin=12pt, itemsep=2pt, topsep=2pt]
        \item \textbf{时间管理：} 建议 **17:00前出发** 或调整至 20:00后游览沙面夜景。
        \item \textbf{顺路整合：} 永庆坊与沙面岛同属荔湾区，建议连续游览。
        \item \textbf{交通替代：} \faSubway\ 若侧重效率，推荐 **地铁3号线** 至珠江新城，比驾车节省约40分钟。
    \end{enumerate}
    
    \vspace{3pt}
    
    % 底部 Tip (使用 faClock 图标)
    \colorbox{color_amapgrey}{\parbox{\dimexpr\linewidth-6pt}{
        \scriptsize 
        \textbf{\faClock\ 小贴士：}珠江夜游建议安排在傍晚，大沙头码头可顺路衔接行程。
    }}
\end{tcolorbox}

\end{tcolorbox}

\subsection{Example 3: Long-distance Travel Planning with Amenities}
\begin{tcolorbox}[
    breakable,
    enhanced,
    colframe=color_amapblue,
    colback=mycolor,
    coltitle=white,
    title = {\textbf{Case Study: Cross-province Self-driving Guide}},
    fonttitle=\bfseries
]

% --- 用户 Query ---
\textbf{Query:} 给我出一份攻略，重庆市区自驾到海口走哪条路？然后的高速路费以及中途休息的话推荐休息的地方\\
\textbf{Intent:} planning\_and\_decision.itinerary\_planning.short\_trip\_itinerary
\tcblower

% --- Step 1: 实体解析 ---
\textbf{Action Step 1: Entity Resolution}
\begin{itemize}[leftmargin=12pt, itemsep=0pt]
  \item \textbf{Tool:} \texttt{map\_search\_places} (called twice)
  \item \textbf{Observation:} 
    \begin{itemize}[leftmargin=10pt, itemsep=0pt]
        \item Origin: Chongqing ($29.562, 106.551$)
        \item Destination: Haikou ($20.046, 110.200$)
    \end{itemize}
\end{itemize}
\vspace{5pt}
% --- Step 2: 路线计算 ---
\textbf{Action Step 2: Route Computation}
\begin{itemize}[leftmargin=12pt, itemsep=0pt]
  \item \textbf{Tool:} \texttt{map\_compute\_routes}
  \item \textbf{Arguments:} \{``origin'': $P_{CQ}$, ``destination'': $P_{HK}$, ``modes'': [``driving'']\}
  \item \textbf{Observation:} 
    \begin{itemize}[leftmargin=10pt]
        \item \textbf{Metrics:} 1381.7 km / 14h 43m / Toll: 819 CNY.
        \item \textbf{Key Roads:} G75 Lanhai Expy, G15 Shenhai Expy.
    \end{itemize}
\end{itemize}
\vspace{5pt}
% --- Step 3: 沿途搜索 ---
\textbf{Action Step 3: Amenities Search Along Route}
\begin{itemize}[leftmargin=12pt, itemsep=0pt]
  \item \textbf{Tool:} \texttt{map\_search\_along\_route}
  \item \textbf{Arguments:} \{``query'': ``加油站,餐厅'', ``origin'': $P_{CQ}$, ``destination'': $P_{HK}$, ``transport\_mode'': ``driving''\}
  \item \textbf{Observation:} Found 3 stops (KFC Drive-thru, Duyun Service Area Buffet, etc.)
\end{itemize}

\vspace{5pt}

% --- Response Generator (模拟最终渲染效果) ---
\textbf{Response Generator (Rendered Output):}
\begin{tcolorbox}[colback=white, colframe=color_amapblue!30!white, arc=4pt, boxrule=1pt, left=4pt, right=4pt, top=4pt, bottom=4pt]
    \small
    % BLUF
    \textbf{BLUF:} 全程约1382公里，需14.5小时，高速费约819元。建议在贵州都匀段进行中途休整。
    
    \par\noindent\rule{\textwidth}{0.4pt}
    
    % 路线概况 (使用 faRoad 图标)
    \textbf{\textcolor{color_amapblue}{\faRoad\ 路线概况}}
    \begin{itemize}[leftmargin=12pt, itemsep=2pt, topsep=2pt]
        \item \textbf{核心指标：} 1381.7 km | 14h 43m | ¥819 (含过路费)
        \item \textbf{主要高速：} G75兰海高速 $\rightarrow$ G7522贵北高速 $\rightarrow$ G15沈海高速
        \item \textbf{关键节点：} G75巴南收费站(起点) ... 徐闻港收费站(终点)
    \end{itemize}
    
    \par\noindent\rule{\textwidth}{0.4pt}
    
    % 休息推荐 (使用 faCoffee 图标)
    \textbf{\textcolor{color_amapblue}{\faCoffee\ 中途休息推荐}}
    \begin{enumerate}[leftmargin=12pt, itemsep=2pt, topsep=2pt]
        \item \textbf{都匀加油站自助餐厅 (贵州段)}
        \begin{itemize}[leftmargin=6pt, label={}, itemsep=0pt]
            \item \scriptsize \textcolor{gray}{位于G75高速路口 | 自助中餐 | 适合长途补给}
        \end{itemize}
        \item \textbf{肯德基车速取 (重庆出发段)}
        \begin{itemize}[leftmargin=6pt, label={}, itemsep=0pt]
            \item \scriptsize \textcolor{gray}{南岸加油站旁 | 西式快餐 | 适合出发前用餐}
        \end{itemize}
    \end{enumerate}
    
    \vspace{3pt}
    
    % 底部 Tip (使用 faShip 图标 - 特指轮渡)
    \colorbox{color_amapgrey}{\parbox{\dimexpr\linewidth-6pt}{
        \scriptsize 
        \textbf{\faShip\ 特别提示：}路线包含 **琼州海峡轮渡** (徐闻港-新海港)，船票需另行购买，建议提前在网上预约过海。
    }}
\end{tcolorbox}

\end{tcolorbox}

\subsection{Example 4: EV Long-distance Route \& Charging Planning}
\begin{tcolorbox}[
    breakable,
    enhanced,
    colframe=color_amapblue,
    colback=mycolor,
    coltitle=white,
    title = {\textbf{Case Study: EV Inter-provincial Travel Guide}},
    fonttitle=\bfseries
]

% --- 用户 Query ---
\textbf{Query:} 从南通开新能源车到张家口万龙滑雪场，续航500公里，求规划行车路线和沿途充电方案\\
\textbf{Intent:} planning\_and\_decision.route\_planning
\tcblower

% --- Step 1: 实体解析 ---
\textbf{Action Step 1: Entity Resolution}
\begin{itemize}[leftmargin=12pt, itemsep=0pt]
  \item \textbf{Tool:} \texttt{map\_search\_places} (x2)
  \item \textbf{Observation:} 
    \begin{itemize}[leftmargin=10pt, itemsep=0pt]
        \item Origin: Nantong City ($31.981, 120.894$)
        \item Dest: Wanlong Ski Resort ($40.963, 115.398$)
    \end{itemize}
\end{itemize}

% --- Step 2: 路线计算 ---
\textbf{Action Step 2: Route Computation}
\begin{itemize}[leftmargin=12pt, itemsep=0pt]
  \item \textbf{Tool:} \texttt{map\_compute\_routes}
  \item \textbf{Arguments:} \{``origin'': $P_{NT}$, ``destination'': $P_{WL}$, ``modes'': [``driving'']\}
  \item \textbf{Observation:} 
    \begin{itemize}[leftmargin=10pt]
        \item \textbf{Metrics:} 1303 km | 14h 55m.
        \item \textbf{Highways:} G15 Shenhai $\rightarrow$ G25 Changshen $\rightarrow$ G2 Jinghu.
    \end{itemize}
\end{itemize}

% --- Step 3: 沿途充电搜索 ---
\textbf{Action Step 3: Charging Infrastructure Search}
\begin{itemize}[leftmargin=12pt, itemsep=0pt]
  \item \textbf{Tool:} \texttt{map\_search\_along\_route}
  \item \textbf{Arguments:} \{``query'': ``充电站'', ``transport\_mode'': ``driving''\}
  \item \textbf{Observation:} Found 50+ stations. Selected strategic stops at ~200km and ~500km intervals (e.g., Binhai SA, Linshu SA).
\end{itemize}

\vspace{5pt}

% --- Response Generator (模拟最终渲染效果) ---
\textbf{Response Generator:}
\begin{tcolorbox}[
    breakable, % 允许跨页
    colback=white, 
    colframe=color_amapblue!30!white, 
    arc=4pt, 
    boxrule=1pt, 
    left=4pt, right=4pt, top=4pt, bottom=4pt
]
    \small
    % BLUF
    \textbf{BLUF:} 全程1305公里，需14小时。基于500km续航，建议 **3次充电**，尤其需确认河北段补能设施。
    
    \par\noindent\rule{\textwidth}{0.4pt}
    
    % 路线概况
    \textbf{\textcolor{color_amapblue}{\faRoute\ 路线概况}}
    \begin{itemize}[leftmargin=12pt, itemsep=2pt, topsep=2pt]
        \item \textbf{核心指标：} 1305 km | 13h 59m | 明日08:08到达
        \item \textbf{主要高速：} G18荣乌 $\rightarrow$ G25长深 $\rightarrow$ S3801京礼
        \item \textbf{拥堵预警：} S15京津高速拥堵1.2km，预计延误10分钟。
    \end{itemize}
    
    \par\noindent\rule{\textwidth}{0.4pt}
    
    % 充电策略 (表格还原)
    \textbf{\textcolor{color_amapblue}{\faChargingStation\ 充电补能规划}}
    \begin{center}
    \renewcommand{\arraystretch}{1.2}
    \resizebox{\linewidth}{!}{
    \begin{tabular}{|c|l|l|l|}
    \hline
    \textbf{\#} & \textbf{地点} & \textbf{距离起点} & \textbf{操作建议} \\ \hline
    1 & 南通理想超充站 & 0 km & \textbf{出发前充满} \\ \hline
    2 & 沂南东服务区 (山东) & $\approx$ 500 km & 充至80\%，避免排队 \\ \hline
    3 & 万龙滑雪场 (河北) & $\approx$ 1300 km & \textbf{需现场确认} \\ \hline
    \end{tabular}
    }
    \end{center}

    % 警告提示 (模拟引用块)
    \begin{tcolorbox}[colback=yellow!10!white, colframe=yellow!50!white, left=2pt, right=2pt, top=2pt, bottom=2pt, boxrule=0.5pt, arc=2pt]
        \scriptsize \textbf{\faExclamationTriangle\ 风险提示：} 工具检索显示 **河北段高速暂无明确充电站**。强烈建议在山东临沂充满电，并提前联系滑雪场（如游客中心/地下停车场）确认目的地充电设施。
    \end{tcolorbox}
    
    \vspace{3pt}
    
    % 下一步建议
    \textbf{\textcolor{color_amapblue}{\faMapSigns\ 下一步行动}}
    \begin{itemize}[leftmargin=12pt, itemsep=0pt, topsep=2pt]
        \item \textbf{出发前：} 下载离线地图，避免高速信号中断。
        \item \textbf{途中：} 每行驶300km检查电量，保持30\%以上余量。
        \item \textbf{抵达：} 导航搜索“万龙滑雪场2号地下停车场”寻找充电桩。
    \end{itemize}

\end{tcolorbox}

\end{tcolorbox}

\end{CJK*} % 结束中文环境

\section{Training Dynamic}
\begin{figure}[htbp]
    \centering
    \includegraphics[width=0.85\textwidth]{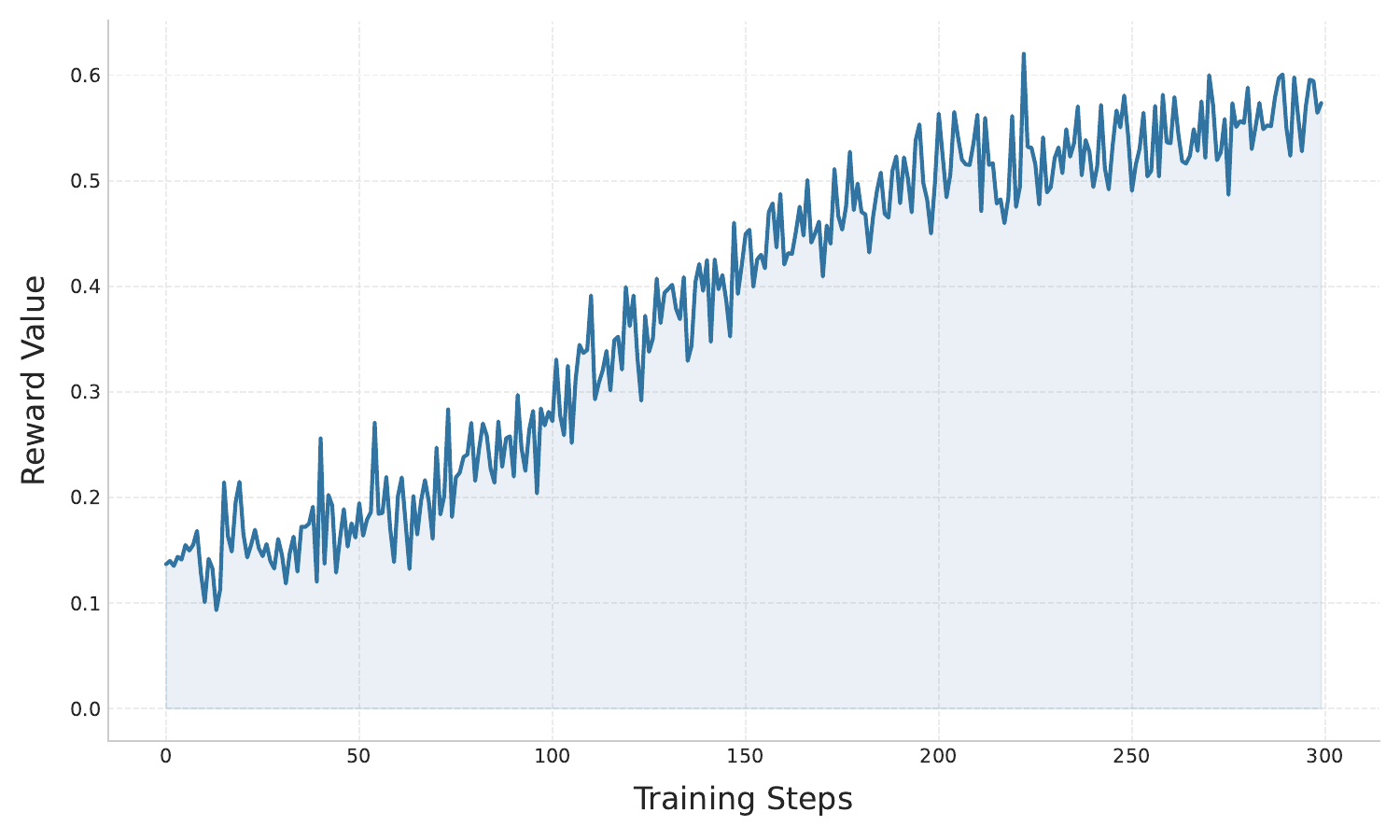} 
    \caption{Training reward.}
    \label{fig:reward_curve}
\end{figure}

Figure~\ref{fig:reward_curve} shows the progression of the reward signal over the process of the training iterations.

\end{document}